\documentclass[10pt,journal,compsoc]{IEEEtran}
\ifCLASSOPTIONcompsoc
  \usepackage[nocompress]{cite}
\else
  \usepackage{cite}
\fi
\ifCLASSINFOpdf
\else
\fi
\usepackage{CJKutf8}

\usepackage{amsthm,amsmath,amssymb,bm}
\usepackage{graphicx}
\usepackage{subfigure}
\usepackage{booktabs}
\usepackage{threeparttable}
\usepackage{multirow}

\usepackage{times}
\usepackage{latexsym}

\usepackage{ragged2e}
\usepackage{xcolor}
\newcommand{\citep}{\cite} 
\newcommand{\citet}{\cite}

\usepackage{url}

\newcommand*{\dif}{\mathop{}\!\mathrm{d}}

\begin{document}
%
\title{Paint4Poem: A Dataset for Artistic Visualization of Classical Chinese Poems}
%
%
%
%

\author{
        Dan Li\IEEEauthorrefmark{1},
        Shuai Wang\IEEEauthorrefmark{1},
        Jie Zou\IEEEauthorrefmark{1}, 
        Chang Tian\IEEEauthorrefmark{2},
        Elisha Nieuwburg\IEEEauthorrefmark{1},
        Fengyuan Sun\IEEEauthorrefmark{1},
        Evangelos Kanoulas\IEEEauthorrefmark{1}\\
    
    \IEEEauthorblockA{\IEEEauthorrefmark{1}Institute of Informatics,
    University of Amsterdam\\
    \IEEEauthorrefmark{2}Catholic University of Leuven\\}

\IEEEcompsocitemizethanks{
\IEEEcompsocthanksitem D. Li, S. Wang, J. Zou, E. Nieuwburg, F. Sun, E. Kanoulas are with the Institute of Informatics, University of Amsterdam, Amsterdam 1098XH, The Netherlands (email: d.li@uva.nl, shuai.wang@student.uva.nl, j.zou@uva.nl, eanieuwburg4@gmail.com, fengyuansun2000@gmail.com, e.kanoulas@uva.nl).\protect\\

\IEEEcompsocthanksitem C. Tian is with Catholic University of Leuven (email: chang.tian@kuleuven.be). }
}

\IEEEtitleabstractindextext{%
\justifying
\begin{abstract}
In this work we propose a new task -- artistic visualization of classical Chinese poems, where the goal is to generate paintings of a certain artistic style for classical Chinese poems. 
For this purpose, we construct a new dataset called Paint4Poem. 
The first part of Paint4Poem consists of 301 high-quality poem-painting pairs collected manually from an influential modern Chinese artist Feng Zikai.
As its small scale poses challenges for effectively training poem-to-painting generation models, we introduce the second part of Paint4Poem, which consists of 3,648 caption-painting pairs collected manually from Feng Zikai's paintings and 89,204 poem-painting pairs collected automatically from the web. We expect the former to help learning the artist painting style as it contains his most paintings, and the latter to help learning the semantic relevance between poems and paintings. 
Further, we analyze Paint4Poem regarding poem diversity, painting style, and the semantic relevance between poems and paintings. 
We create a benchmark for Paint4Poem: we train two representative text-to-image generation models -- AttnGAN and MirrorGAN, and evaluate their performance regarding painting pictorial quality, painting stylistic relevance, and semantic relevance between poems and paintings. 
The results indicate that the models are able to generate paintings that have good pictorial quality and mimic Feng Zikai's style, but the reflection of poem semantics is limited.
The dataset also poses many interesting research directions on this task, including transfer learning, few-shot learning, text-to-image generation for low-resource data etc. 
The dataset is publicly available. (\url{https://github.com/paint4poem/paint4poem})

\end{abstract}

}

\maketitle

\IEEEdisplaynontitleabstractindextext

%
\IEEEpeerreviewmaketitle

\IEEEraisesectionheading{\section{Introduction}
\label{sec:introduction}}

%
%
%
%


\begin{figure*}[!htb]
    \centering
    \includegraphics[width=.7\textwidth]{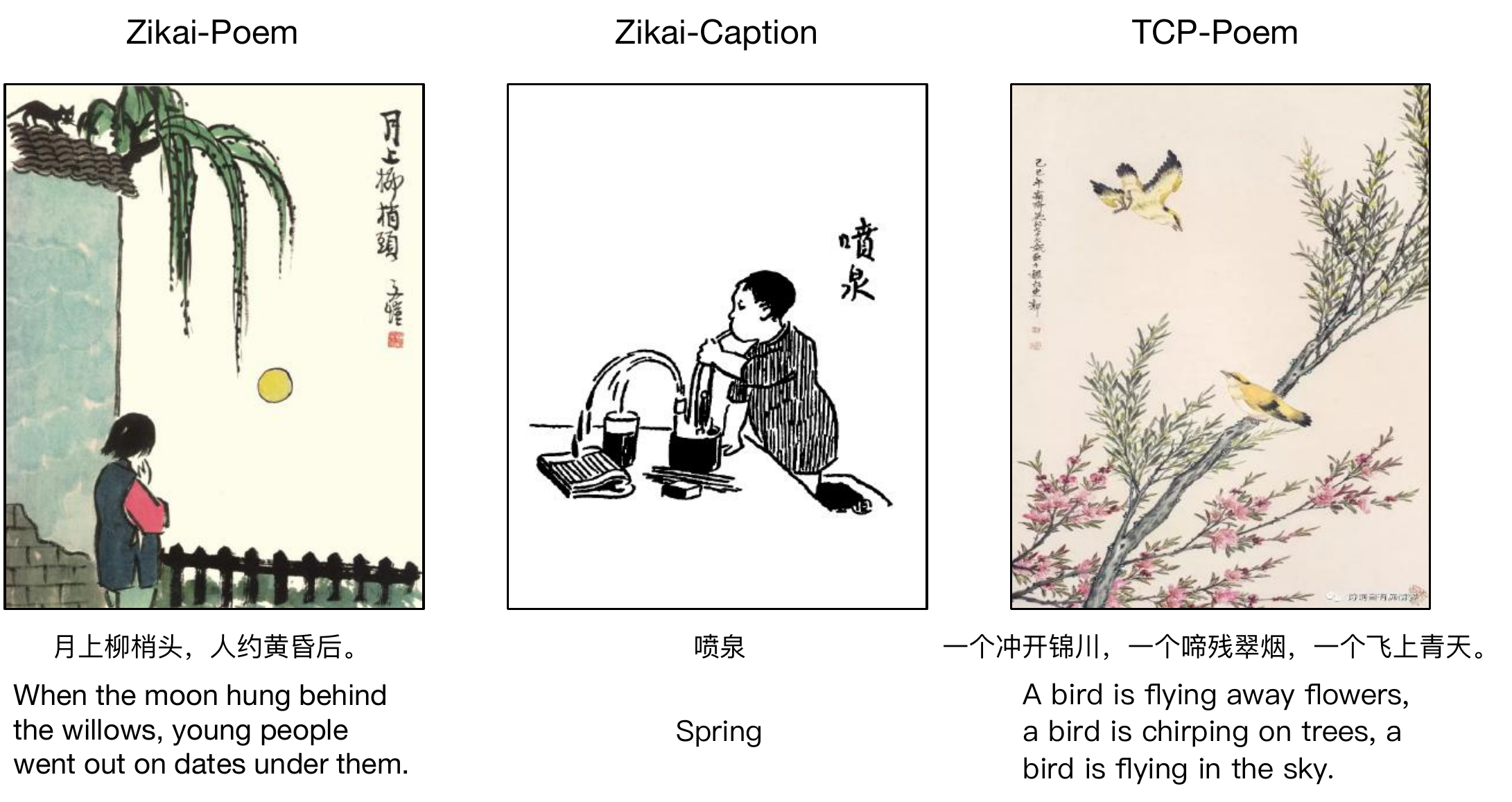}
    \caption{Three examples from the Zikai-Poem, Zikai-Caption, and TCP-Poem datasets. Each example consists of a piece of text and an image describing the text semantics and some other meta data.}
    \label{fig:paint4poem-example}
\end{figure*}

\begin{table*}[!htbp]
\centering
\caption{An overview of the Paint4Poem dataset. }
\label{tbl:data-statistics}
\begin{tabular}{cccc}
\toprule 
 Dataset & Zikai-Poem & Zikai-Caption  & TCP-Poem \\ \midrule
Example number & 301       & 3,648      & 89,204  \\
For training or test  & Train (256) \& test (75)  & Train  & Train \\
Text type             &  Poem   &  Caption  & Poem  \\ 
Image style         &  Feng Zikai style &  Feng Zikai style & Traditional Chinese painting style  \\
Text image matching annotation & Manual  & Manual  & Automatic \\
\bottomrule
\end{tabular}
\end{table*}

\IEEEPARstart{C}{lassical} Chinese poetry is a precious cultural heritage. 
It dates back to the early first millennium BC and develops as one of the largest and longest continuous literature in the world~\citep{watson1984columbia}. 
It has great aesthetic, academic and social value as it enables people to appreciate and study the language, as well as learn the culture.
However, it can be difficult to understand because poetry is regarded as the art of language and beauty and characterized by creative use of language and high abstraction of meaning. 
On the other hand, the use of paintings to visualize classical Chinese poems is a tradition.
There is a concept of convertibility between classical Chinese poetry and Chinese paintings, meaning that they are parallel and comparable~\cite{frankel1957poetry}.  
For example, the Chinese poet Su Shih (A.D. 1036-1101) comments that ``Tufu's poems are figureless paintings, Han Kan's paintings are wordless poems''. (Tufu is a poet and Han Kan is a painter.)
Paintings make poems easy to understand via the art of vision   and they also create new perspectives for the original poems.
Moreover, the effort of bringing together classical Chinese poems and paintings enriches the development of art and creativity for language and vision.  

On the other hand, machine learning based art and creativity continues to grow and attract a wider audience to machine learning researchers and artists~\cite{ai4artworkshop2017, ai4artworkshop2018, ai4artworkshop2019, ai4artworkshop2020, ai4artworkshop2021}. 
There have been a boost of work, for example, using generative models on new types of media creation across language and images, including recent advances such as the GPT model~\cite{radford2019language}, the CLIP model~\cite{radford2021learning} and the DALLE model~\cite{ramesh2021zero}.
The media created automatically or with human interaction by these models have been innovating the development of art and creativity and have broad sociocultural impact and attracts discussions from artists~\cite{ai4artworkshop2017, ai4artworkshop2018, ai4artworkshop2019, ai4artworkshop2020, ai4artworkshop2021}. 

As far as we know, there has been no work bridging classical Chinese poems and artistic painting generation.
In this work, we are interested in studying machine learning based painting generation for classical Chinese poems, which we denote \textit{poem-to-painting} task.
We are specifically interested in the paintings of Feng Zikai, who is an influential modern Chinese artist. Many of his paintings are inspired by classical Chinese poems and created new perspectives of understanding poems~\citep{barme2002artistic}. 
Although Feng Zikai's poem-inspired paintings have great aesthetic value, the amount is limited because he has passed away. The study on painting generation for classical Chinese poems will bring both his understanding of classical Chinese poems and his painting techniques back to life. It can be also generalized to other artists and attracts discussions in both the machine learning and art communities.

To the best of our knowledge, there is no suitable dataset for the poem-to-painting task. We thus construct the \textit{Paint4Poem} dataset. 
The first part consists of 301 high-quality poem-painting pairs collected manually from Feng Zikai's paintings, which serves as both the training and test set. We name it \textit{Zikai-Poem}.
As its small scale poses a big challenge for training, we introduce the second part, which consists of 3,648  caption-painting pairs collected manually from Feng Zikai's painting work (named \textit{Zikai-Caption}) and 89,204 poem-painting pairs of traditional Chinese painting (TCP) style collected automatically from the web (named \textit{TCP-Poem}). 
Zikai-Caption contains most paintings from Feng Zikai, but the captions involved are short text, typically 2 or 3 words. We expect it to help learning the artist painting style.
TCP-Poem contains a much larger number of poems and paintings of traditional Chinese painting style. The poems and paintings are automatically collected  from the web using a scraping script and then paired using an image style classification model. We expect it to help learning text image alignment. 
To sum up, the Paint4Poem is the first dataset for artistic visualization of classical Chinese poems. 

Paint4Poem poses several new research challenges to the community of machine learning based art and creativity.
Although it is possible to apply existing text-to-image generation models generate stylistic paintings for classical Chinese poems, there are new technical challenges such as poem representation learning, painting representation learning, aligning of poem and painting representations, stylizing of generated paintings. There are also challenges regarding evaluation of the models. Except for generating paintings that reflect the semantics of poems and the painting style of the artist, creativity is also an important criteria and it is unclear how to evaluate the creativity yet~\cite{franceschelli2021creativity}.

To assess challenges that the Paint4Poem dataset poses to current visual generative models, we create a benchmark. 
We test the dataset against two state-of-the-art text-to-image generation models, including AttnGAN~\citep{Tao18attngan} and MirrorGAN~\citep{qiao2019mirrorgan}. 
To understand how to best utilize the dataset, we also test transfer learning techniques in terms of improving model performance, i.e. using the two different auxiliary datasets Zikai-Caption or TCP-Poem for pretraining and using Zikai-Poem training set for fine-tuning. 
For fair comparison, all the models are tested on the test set of Zikai-Poem. 
We are interested in evaluating whether the models are able to generate paintings that have good pictorial quality, represent Feng Zikai's style, and reflect the semantic meaning of the given poems. 
Inspired by previous work on the evaluation of image generation~\cite{Tao18attngan, wang2021evaluate}, we use \textit{inception score} (IS), \textit{precision@1} (P@1), and \textit{global effects} \& \textit{local patterns} (GE \& LP) to evaluate the aforementioned models in terms of the three aspects.
We also conducted human evaluation to understand how difficult the task is for human and which models generate better paintings.

We are interested in the performance of different benchmark models on Zikai-Poem, the impact of transfer learning with the two different auxiliary datasets Zikai-Caption and TCP-Poem.
%
First, MirrorGAN performs better than AttnGAN on IS and P@1 scores and they performs on par with each other on GE and LP scores. The generated paintings by both models have acceptable pictorial quality and stylistic relevance and still have large improvement space, but their semantic relevance is limited due to the difficulty of learning poem and painting alignment.
Second, the strategy of pretraining on Zikai-Caption and fine-tuning on Zikai-Poem helps to improve IS and GE \& LP scores with a big margin compared with purely training on Zikai-Poem, which is the same as our expectation on improving pictorial quality and stylistic relevance; while the strategy of pretraining on TCP-Poem and fine-tuning on Zikai-Poem improves all the four metrics, though with a marginal improvement. 
Finally, human evaluation shows that the generated paintings from AttnGAN are better than MirrorGAN, which conflicts with the result of IS and P@1 and further improvement regarding automatic evaluation should be conducted; and the task of generating painting given poems is difficult.

To sum up, we demonstrated  the possibility of visualizing classical Chinese poems in desired painting styles. Our contributions are:
\begin{itemize}
\item  A new task of generating artistic paintings for classical Chinese poems.
\item  A new dataset of Paint4Poem and a thorough analysis of the dataset with regard to poem diversity, painting style, and the semantic relevance between  poems and paintings.
\item  A benchmark that contains one test set, three training sets, two state-of-the-art models, and four evaluation metrics. 
\item  Empirical experiments on how the state-of-the-art models perform and how different transfer learning strategies impact model performance.
\end{itemize}

\section{Dataset Collection}
In this section we explain the pipeline for collecting our dataset, which  consists of the Zikai-Poem dataset, the Zikai-Caption dataset and the TCP-Poem dataset.
The Zikai-Poem dataset plays a key role in training and evaluating poem-to-painting generation models. It covers famous paintings from Feng Zikai and each painting is paired with a poem and abundant auxiliary information.  The Zikai-Poem dataset contains 301 examples in total, posing research challenges on few-shot learning and transfer learning.  

To encourage researches on such challenges,  we provide two related datasets:  Zikai-Caption and TCP-Poem. 
The Zikai-Caption dataset covers 3,648 Feng Zikai's paintings, which accounts for most of his less-famous paintings. Each painting is accompanied with a short caption, typically two or three words. This dataset can be used to learn Feng Zikai's painting style.

The scale of the Zikai-Caption and TCP-Poem datasets is rather small compared with mainstream text-to-image dataset such as MS COCO~\cite{lin2014microsoft}, therefore we automatically construct a large-scale but noisy dataset -- TCP-Poem. Starting from a large poem corpus we scrape for each poem 10 relevant images of traditional Chinese paintings using search engine, and then train a binary image style classification model to keep 1 image that are most possibly in traditional Chinese painting style. This results in collecting 89,204 automatically matched poem-painting pairs.
It can be used to learn poem representation, painting representation and the alignment of poem representation and painting representation.

The overall statistics of the dataset are summarized in Table~\ref{tbl:data-statistics}.

\subsection{Collection of Zikai-Poem}
The Zikai-Poem dataset plays a key role in training and evaluating models that visualize poems in Feng Zikai's style. 
We collect data from two books~\citep{gushici, gushiwen} that contain Feng Zikai's paintings. 
The paintings in the books are colored and have good quality among all Feng Zikai's paintings, thus they can be regarded as good representatives of his painting style. 
Each painting is associated with a poem as its theme, an explanation in modern Chinese, and other commentary texts providing necessary background information to understand the poem. 

The resource results in 301 poem-painting pairs.
Each example in the dataset contains the following metadata: PoemID, PoemText, PoemTitle, PoemDynasty, PoemAuthor, Explanation, Commentary, and PaintingID.
The metadata respectively describe: 
the poem identification number, 
the text of the poem, 
the poem title, 
the dynasty the poem is written in, 
the author of the poem, 
notes on unusual words, 
the poem commentary, 
and the painting identification number.


\subsection{Collection of Zikai-Caption}
As Zikai-Poem is too small to effectively train and evaluate poem-to-painting generation models, we provide an auxiliary dataset which is suboptimal for the poem-to-painting task but has larger number of examples than Zikai-Poem.

We collect data from various books~\citep{monochrome, comics} and a website~\citep{classic_comics}. These resources cover most of Feng Zikai's less-famous paintings compared with Zikai-Poem. Most paintings are monochrome and accompanied with very limited textual information, typically a \textit{poem segment} or two or three words as their captions.  We believe these paintings are helpful to learn Feng Zikai's painting style.

Note that the books are scanned PDF files. The texts are extracted by using an OCR application and the paintings are gathered by using an online PDF-to-image converter. The paintings are cropped in order to remove all unnecessary information present on them. Finally, the texts and paintings are checked manually and those deem unfit for poem-to-painting generation are removed from the dataset.

The resource results in 3648 caption-poem pairs.
Each example in the dataset contains the following metadata: PoemID, PaintingCaption, and PaintingID.
This metadata respectively describe: 
the poem identification number, 
the caption of the painting, 
and the painting identification number.

\begin{figure}[!htbp]
    \centering
    \includegraphics[width=0.5\textwidth]{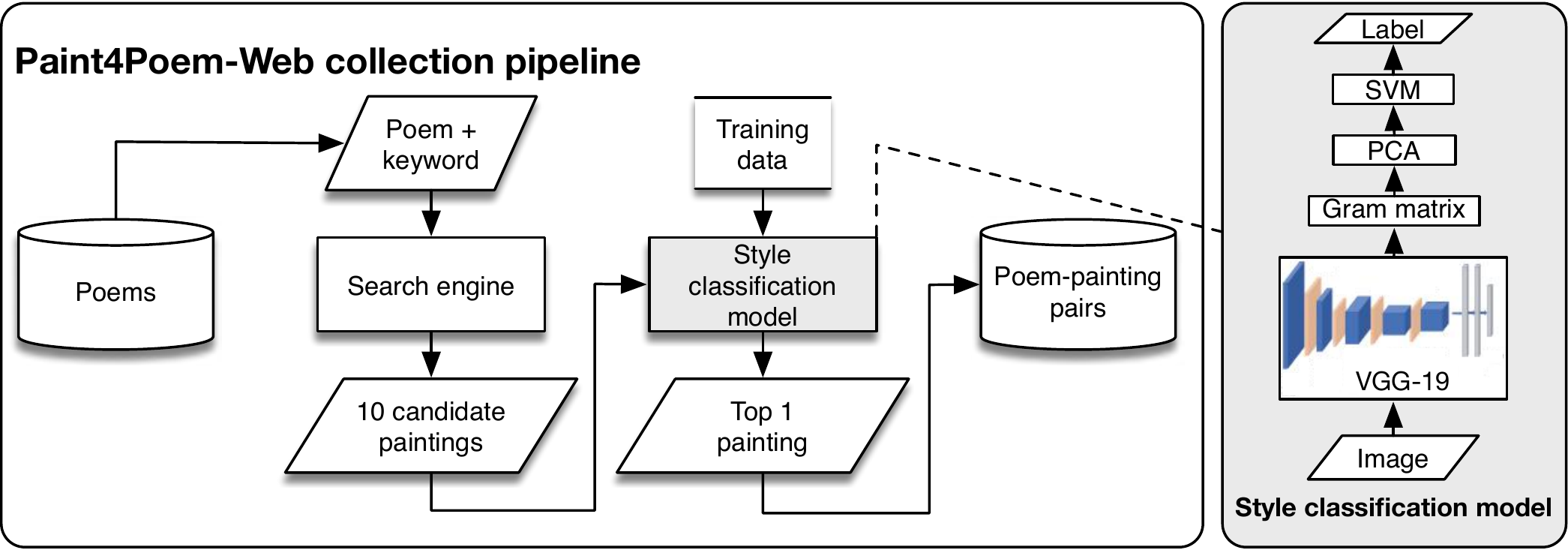}
    \caption{Data collection pipeline for TCP-Poem.}
    \label{fig:pipeline}
\end{figure}

\subsection{Collection of TCP-Poem }
\label{sec:Paint4Poem-Web}
The second auxiliary dataset to augment the small Zikai-Poem dataset is TCP-Poem, which provides a very large-scale paintings of traditional Chinese painting style paired with poems. We include traditional Chinese paintings because Feng Zikai's paintings is considered belonging to this type. 
We first collect large-scale poems from an open-source repository and scrape relevant images of traditional Chinese paintings from the web, and then train a binary image style classification model to filter out images that are not in traditional Chinese painting style. 
This resulted in collecting 89,204 automatically matched poem-painting pairs.
The data collection pipeline is illustrated in Fig.~\ref{fig:pipeline}. 

\textbf{Poem resources.}
Two sources are used to collect poems. The first is a Chinese poem indexing site\footnote{www.gushiwen.org} containing well-known poem segments. The assumption is that well-known poems are popular theme for artists to create paintings and thus are more likely to retrieve relevant images at the next stage. It results in 6,152 poem segments. The second is an open-source repository\footnote{https://github.com/Werneror/Poetry} covering most classical Chinese poems in literature. It results in 83,052 poem segments. Such a high-volume training data is beneficial for poem representation learning. 

\textbf{Candidate images retrieval.}
We use web search engines to retrieve a set of candidate images for each poem, given that modern search engines provide effective and efficient image search services and they can query high volumes of images. 
We define an image to be \textit{stylistically relevant} if it represents a painting in the traditional Chinese painting style, and \textit{semantically relevant} if it portrays the meaning of that poem.
We conduct a trial experiment to find the best search engine and the best way of forming search queries in terms of retrieving semantically relevant images.  We choose the Baidu\footnote{www.baidu.com} search engine and use a combination of the poem segment and the keyword `\begin{CJK*}{UTF8}{gbsn}传统国画\end{CJK*}', which translates to `traditional Chinese painting', as the search queries. The top-10 images returned from the search engine are downloaded using a web scraping script.

\textbf{Image style classification.}
It is reasonable to assume that the candidate images are semantically relevant to the poem query because this is the designing goal of a search engine. But the candidate images are not necessarily stylistically relevant to the poem query, therefore we need to filter out these images. We form the task as an image style classification task. For each poem and its 10 candidate images, only the image with the highest predicted probability by our image style classifier is paired with the poem.  

The image style classifier consists of a pre-trained VGG-19 network to extract image vectors~\citep{simonyan2014very}, a Gram matrix layer to extract style vectors~\citep{gatys2015neural}, a principal component analysis model to reduce vector dimension~\citep{tipping1999probabilistic}, and a support vector machine classifier~\citep{suykens1999least} to predict labels. The loss is a binary cross-entropy loss. 

We now describe the training and test set. To develop a training set, for positive examples, we use 173 paintings from a dataset of traditional Chinese paintings~\citep{chinese-painting-dataset} to save labelling effort; we also manually label 173 positive paintings from the candidate painting collection. The negative examples are 346 negative paintings manually labelled from the candidate painting collection. The strategy for the test set is the same, with 77 positive and 77 negative examples.
The trained classifier achieves an accuracy of 96.7\% on the test set.  The good performance guarantees that the paintings of the TCP-Poem dataset are stylistically consistent.

\section{Dataset Analysis}

\subsection{Zikai-Poem}

\textbf{Poems diversity.}
The poems in the Zikai-Poem dataset cover 162 poets from all ancient Chinese dynasties, 
indicating a writing style diversity. Furthermore, it is interesting to present \textit{imagery} diversity. 
Imagery, which uses a certain object to create a picture in readers' mind and evoke emotions, is key to understand classical Chinese poetry. It is the imageries that inspire artists to create paintings based on poems.
We use the imagery taxonomy in \textit{ShiXueHanYing} (\textit{Shi})~\citep{shixuehanying}. It consists of 1024 imageries along with the corresponding synonyms that are often used in classical Chinese poetry. We use \textit{exact match} to count imagery occurrence in the poems.
The dataset covers 82\% of the \textit{Shi} imageries. Table \ref{tbl:zikai-poem-diversity} lists the ten most frequent \textit{Shi} imageries in the dataset. 

\begin{table}[!htb]
\centering
\caption{The ten most occurring \textit{Shi} imageries in the Zikai-Poem dataset, with their corresponding English translations and the number of poems in which the imagery occurs.}
\label{tbl:zikai-poem-diversity}
\begin{tabular}{lll}
\toprule
Topic & English translation & \# Poem \\ \midrule
\begin{CJK*}{UTF8}{gbsn}日\end{CJK*} & sun & 216\\
\begin{CJK*}{UTF8}{gbsn}风\end{CJK*} & wind, breeze &  116 \\
\begin{CJK*}{UTF8}{gbsn}花\end{CJK*} & flower, blossom, bloom & 113 \\
\begin{CJK*}{UTF8}{gbsn}春\end{CJK*} & spring, love & 98 \\
\begin{CJK*}{UTF8}{gbsn}山\end{CJK*} & hill, mountain & 90 \\
\begin{CJK*}{UTF8}{gbsn}水\end{CJK*} & water, river & 82 \\
\begin{CJK*}{UTF8}{gbsn}月\end{CJK*} & moon, month & 76 \\
\begin{CJK*}{UTF8}{gbsn}天\end{CJK*} & sky, heaven, day & 73 \\
\begin{CJK*}{UTF8}{gbsn}美人\end{CJK*} & beautiful woman, beauty & 70 \\
\bottomrule
\end{tabular}
\end{table}

\begin{figure}[!htbp]
    \centering
    \includegraphics[width=.35\textwidth]{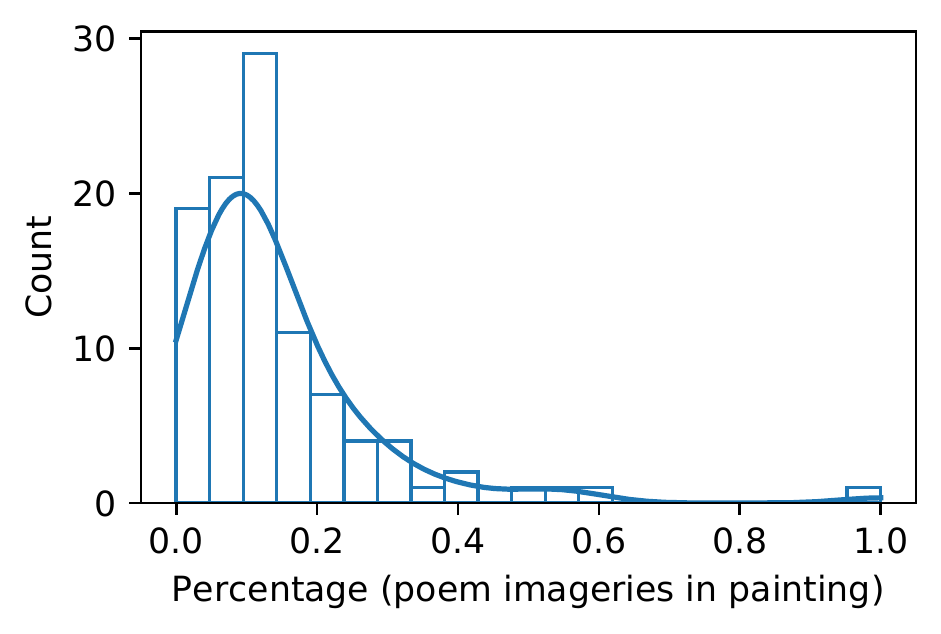}
    \caption{Histogram on the percentage of poem imageries presented in paintings.}
    \label{fig:perct-poem-topic-in-painting}
\end{figure}

\textbf{Painting style.}
The paintings in Fig. \ref{fig:paint4poem-example} are examples of Feng Zikai from which readers can see his painting style.
From a view of artwork criticism, the painting style of Feng Zikai is ``succinct stroking (using Chinese Calligraphy brush), restrained coloring, and empty-emphasizing composition"~\citep[page 9]{gushici}.

In order to have an understanding of the style difference within the Zikai-Poem dataset, we randomly sample 1 painting from Zikai-Poem as style reference and 20 different paintings, and calculate their GE \& LP scores, which aims to evaluate how similar the paintings from the two datasets are in terms of image style. We repeat the process 5 times.
In Table~\ref{tbl:style-similarity}, the GE \& LP scores are 0.85 and 0.58. Existing study has shown that several state-of-the-art style transfer models can achieve GE scores ranging from 0.6 to 0.9 and LP scores ranging from 0.3 to 0.6~\citet{wang2021evaluate}.  It indicates that the style of the paintings are consistent in term of both global effect and local pattern.

\textbf{Semantic relevancy of poem-painting pairs.}
We manually label the poems and paintings with the \textit{Shi} imageries. Then we compute the percentage of poem imageries that are actually portrayed in the corresponding paintings.
Fig.~\ref{fig:perct-poem-topic-in-painting} shows that for 90\% paintings only less than 20\% imageries of the poems are presented in the paintings. 
In short, the diversity of poem imageries are not necessarily reflected in Feng Zikai's paintings, which  challenges the effective training of poem-to-painting generation models.

On the other hand, it is interesting to find that Feng Zikai prefers repeatedly portraying typical imageries in his paintings. For example, objects like sun, moon, hill, water, willow, flower, human repeatedly occur in his many different paintings.  
See examples in Fig. \ref{fig:paint4poem-example}. 
The highly-reoccurring patterns in his paintings makes the training of painting generation models easy.


\begin{table}[!htbp]
\centering
\caption{Image style similarity between Zikai-Poem and Zikai-Poem/Zikai-Caption/TCP-Poem.}
\label{tbl:style-similarity}
\begin{tabular}{ccc}
\toprule 
          & \multicolumn{2}{c}{Zikai-Poem}          \\  \cmidrule(lr){2-3}
          & GE & LP         \\  \midrule
Zikai-Poem   & 0.83 & 0.58 \\
Zikai-Caption& 0.64 & 0.57 \\ 
TCP-Poem     & 0.45 & 0.49 \\ 

\bottomrule
\end{tabular}
\end{table}

\begin{figure}[!htbp]
    \centering
    \subfigure[GE]{
   \label{fig:model-likelihood}
        \includegraphics[width=.45\textwidth]{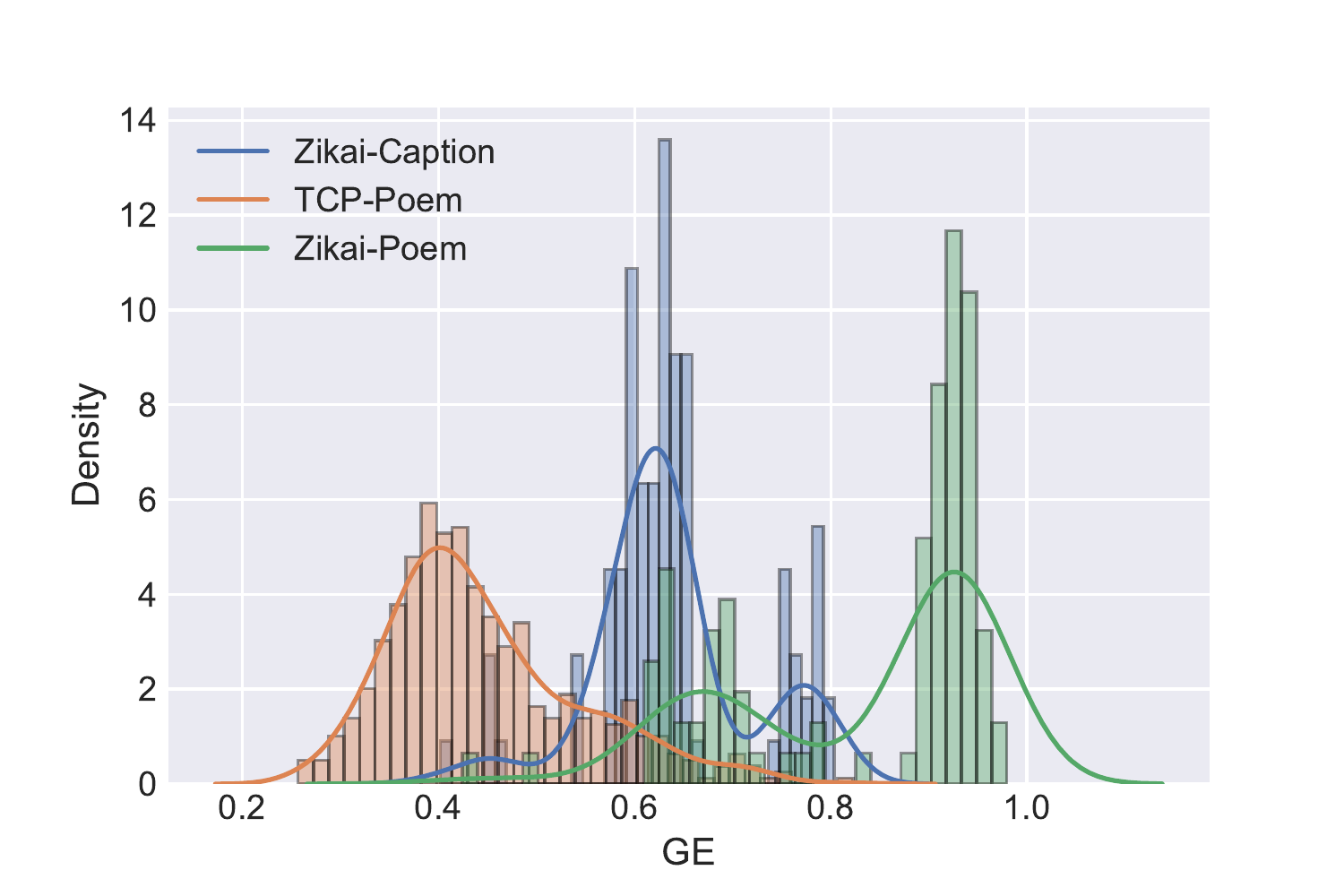}
    }
    \subfigure[LP]{
   \label{fig:model-likelihood}
        \includegraphics[width=.45\textwidth]{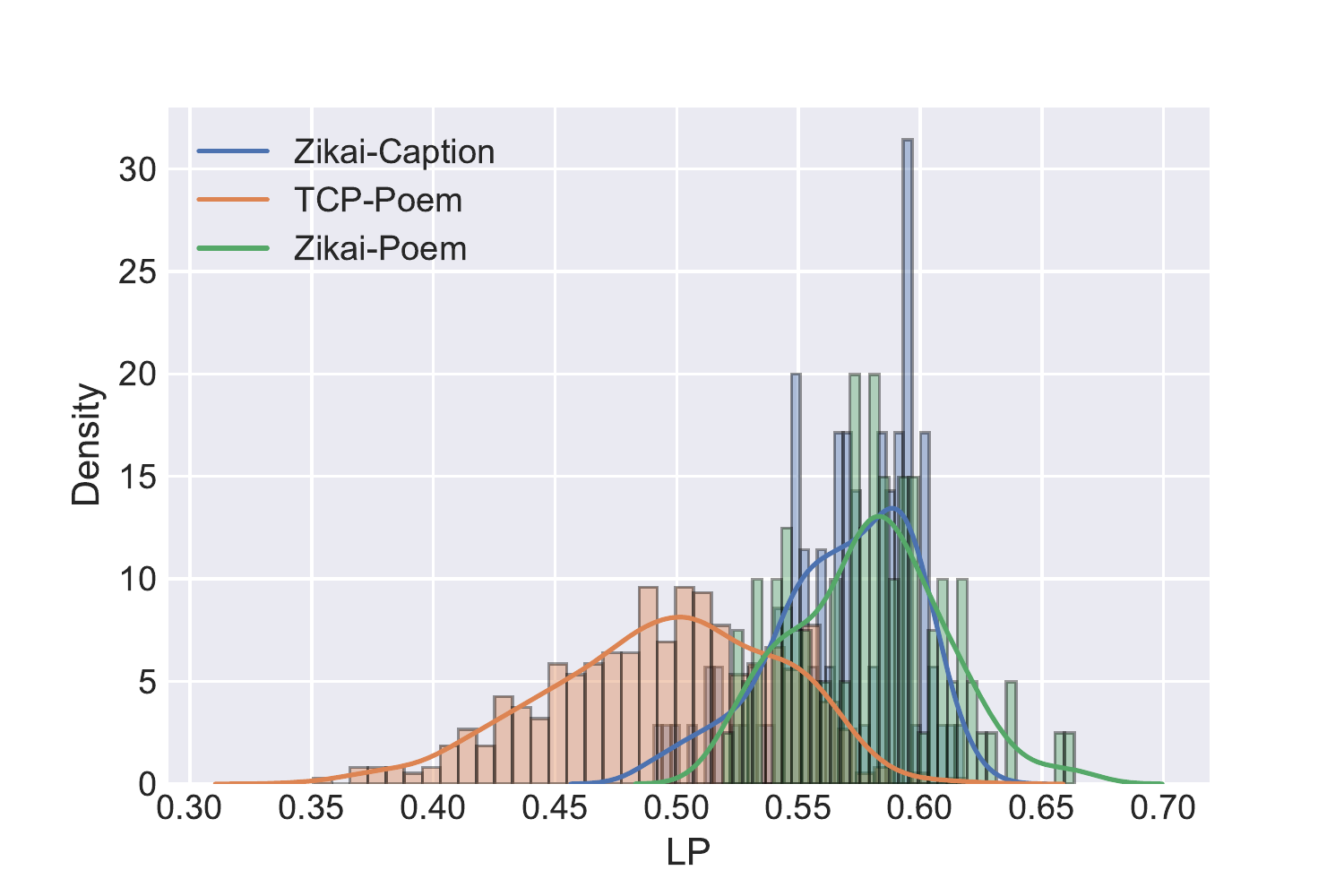}
    }
    \caption{Distribution of image style similarity between Zikai-Poem and Zikai-Poem/Zikai-Caption/TCP-Poem.}
    \label{fig:style-similarity}
\end{figure}

\subsection{Zikai-Caption}

\textbf{Captions.}
The texts in Zikai-Caption are not purely poems, instead they are mixture of short titles, poem segments, and short sentences of modern Chinese that summarize the episodes in the paintings. 
The average length of captions are only 6 characters. 

\textbf{Painting style.}
All the paintings in Zikai-Caption are from Feng Zikai, thus they can provide useful signals to learn his painting style. 
Meanwhile, there are several differences between Zikai-Caption and Zikai-Poem.
For example, only 706 out of 3648 examples are well painted and colored paintings, the rest are simple monochrome paintings; the stroking of paintings is more coarse than that in Zikai-Poem.
In order to have an understanding of the style difference between Zikai-Caption and Zikai-Poem, we randomly sample 1 painting from Zikai-Poem as style reference and 20 paintings from Caption-Poem, and calculate their GE \& LP scores. We repeat the process 5 times.
In Table~\ref{tbl:style-similarity}, the averaged GE \& LP scores between Zikai-Poem and Zikai-Caption are 0.64 and 0.57. In Fig.~\ref{fig:style-similarity}, we find the distributions of GE and LP between Zikai-Caption and Zikai-Poem are close. It indicates the paintings in Zikai-Caption are stylistically similar to Zikai-Poem.

\textbf{Semantic relevancy of caption-painting pairs.}
The captions are from Feng Zikai and thus there is no noise in terms of matching captions with paintings. 
However, the captions should not be understood as a straightforward and literal description of the paintings, instead, they are highly abstract summary of the paintings. 
Although it is a big challenge to learn the matching between texts and images using this dataset, the images alone are helpful for GAN-based models to learn to generate painting of Feng Zikai's style.

\subsection{TCP-Poem}

\textbf{Poems diversity.}
The poems in TCP-Poem are selected across different dynasty periods. A small portion of 6,152 accounts for famous poems and the rest 83,052  are less famous ones.
The large number of poems ensures effective learning of poem representations.

\textbf{Painting style.}
As the large TCP-Poem collection is constructed automatically, not all paintings are in the target painting style. To measure the level of noise with regard to painting style, human evaluation is carried out by sampling 400 poem-painting pairs and labelling their painting style as either traditional Chinese painting style or not. 
The definition on whether a painting is of traditional Chinese painting style or not is the same as that of the positive and negative examples in the training data of the image style classification model introduced in Section~\ref{sec:Paint4Poem-Web}.
It is found that about 85\% of the paintings are of Chinese painting style. 
We conclude that TCP-Poem well represents the traditional Chinese painting style.

In order to have an understanding of the style difference between TCP-Poem and Zikai-Poem, we randomly sample 1 painting from Zikai-Poem as style reference and 100 images from TCP-Poem and calculate their GE \& LP scores. We repeat the process 5 times.
In Table~\ref{tbl:style-similarity}, the averaged GE \& LP scores between Zikai-Poem and TCP-Poem are 0.45 and 0.49, which is lower than that on Zikai-Caption. In Fig.~\ref{fig:style-similarity}, the distribution between TCP-Poem and Zikai-Poem are relatively different.

\textbf{Semantic relevancy of poem-painting pairs.}
To evaluate whether a painting portrays the semantic meaning of its paired poem, we manually sample and analyze 24 poem-painting pairs. 
Only for 5 out of 24 pairs, the paintings are semantically relevant to the the poems. The poems of these positive pairs all contain a content word which is accurately represented in the painting. This finding suggests that the paintings in this dataset are generally not semantically relevant with the corresponding poems.
The main reason is that there do not exist paintings themed on unpopular poems and thus it is not possible to retrieve them.  
Even so, TCP-Poem is still a good collection of classical Chinese paintings as analyzed above.

\section{Benchmarks and Analysis}
\subsection{Research Questions}
The overall question we want to ask is ``\textit{can we generating good paintings of a target style for classical Chinese poems} ''.
To be specific, we are going to answer the following research questions.
\begin{itemize}
\item [RQ1]  How do state-of-the-art text-to-image generation models perform on the dataset?
\item [RQ2]  How to apply transfer learning techniques on the two different auxiliary datasets -- Zaikai-Caption and TCP-Poem to improve model performance?
\item [RQ3]  How do model hyperparameters affect model performance?
\end{itemize}

\subsection{Training and test sets}
We split the Zikai-Poem dataset into training set and test set, the training set contains 3/4 of the examples which are 256 poem-painting pairs and the test set contains 1/4 of the examples which are 75 poem-painting pairs. 
To investigate the performance of state-of-art text-to-image generation models on Zikai-Poem,  we train these models on the training set from scratch and evaluate the performance on the test set (see Section \ref{sec:performance-benchmark-models}).
To investigate the benefit of transfer learning on Zikai-Caption and TCP-Poem, we train benchmark models on these two datasets from scratch and fine-tune on the training set of Zikai-Poem, and then evaluate the performance on the test set of Zikai-Poem (see Section\ref{sec:transfer-learning}).
The statistics of the training sets and the test set are shown in Table~\ref{tbl:data-statistics}. 


\subsection{Evaluation Criteria}
\label{sec:criteria}
Existing work on image generation has not achieved a consensus on evaluation criteria and the criteria mostly focus on the concept of ``realistic''~\citep{huang2018introduction}, or ``visual quality"~\citep{zhang2017stackgan, zhu2019dm}, ``visual fidelity"~\citep{park2019semantic}. When conducting human evaluation, this criterion is too vague and leads to very different evaluation results across human assessors~\citep{salimans2016improved}.
It is also different from our goal of generating stylistic paintings.
We replace the criterion ``realistic" with three criteria designed for the new task in this work. It can serve as a guideline for human evaluation or new evaluation metric design.
The three criteria are  \textit{pictorial quality},  \textit{stylistic relevance}, \textit{semantic relevance}.
\textit{Pictorial quality} means that the objects in the paintings should be easily recognizable from their outlines and colors. From the view of image generation modelling, this property is possible to be guaranteed because existing image encoding models such as convolutional networks~\cite{simonyan2014very} can capture related features through its first few layers. This criterion is partially involved in the concept of ``realistic"~\cite{huang2018introduction}.
\textit{Stylistic relevance} means that the generated paintings should reflect the target style of the training data. Style of paintings refers to ``the distinctive visual elements, techniques, and methods that typify an individual artist's work''~\cite{wiki:paintingstyle}. For example, the painting style of Feng Zikai is recognized to be ``succinct stroking (using Chinese Calligraphy brush), restrained coloring, and empty-emphasizing composition"~\cite[page 9]{gushici}.
\textit{Semantic relevance} means that the generated paintings should reflect imageries in the given poems.

\subsection{Evaluation Metrics}
\label{sec:eval-metrics}
Following existing work on text-to-image generation task~\cite{Tao18attngan, qiao2019mirrorgan, wang2021evaluate}, we use inception score (IS), precision@1 (P@1), and global effects \& local patterns (GE \& LP) as the quantitative evaluation metrics for the three criteria aforementioned.

The inception score measures the quality and variety of images~\cite{salimans2016improved}. 
The Inception model~\cite{szegedy2016rethinking} is first applied to every generated image to get a 1000-dimensional label distribution $p\left(y \mid \bm{x}\right)$, where $\bm{x}$  is the representation of the generated image and $y$ is the label. 
Images with meaningful objects should have low entropy. Besides, the generated images are expected varied, which means the marginal distribution $p\left(y\right) = \int p\left(y \mid \bm{x}\right) \dif \bm{x}$ should have high entropy. 
Finally, the inception score is defined as  
\begin{eqnarray}
\label{eq:is}
IS = \exp \left( \mathbb{E}_{\bm{x}} \mathcal{KL} \left( p\left( y \mid \bm{x} \right) ||\, p\left( y \right) \right) \right).
\end{eqnarray}
Note $IS \in (1, + \infty)$, a large value indicates good performance.

Precision@1 is used to evaluate the visual-semantic similarity with the given text description~\cite{Tao18attngan}.  The evaluation process is modelled as a ranking task: give the generated image and 100 candidate texts which consists of its ground truth text and 9 random mismatched texts, we ranking the 10 candidate texts by calculating the cosine similarity between the text vector and the image vector\footnote{Note that \cite{Tao18attngan} uses 10 candidates; we use 10 candidates because our test set only contains 75 examples.}. If the ground truth text is ranked at top 1, the generated image is considered to be semantically relevant to the ground truth text. $P@1 \in [0, 1]$, a large value indicates good performance.

Global effects  and local patterns are used to evaluate image style quality~\cite{wang2021evaluate}. According to their user study, humans' visual perception on image style is mainly determined by the colors and textures of images. The GE metric considers the color features and texture features between the generated image and the ground truth image, which is defined as 
\begin{eqnarray}
\label{eq:ge}
GE(\bm{x}, \bm{s})=
\frac{1}{2} \cdot
\frac{1}{3} \sum_{c=1}^{3} \frac{\mathcal{H}_{c}(\bm{x}) \cdot \mathcal{H}_{c}(\bm{s})}{\| \mathcal{H}_{c}(\bm{x})\|\cdot\| \mathcal{H}_{c}(\bm{s}) \|} 
+  \\ \nonumber
\frac{1}{2} \cdot
\frac{1}{N} \sum_{l=1}^{N} \frac{\mathcal{G}\left(f_{l}(\bm{x})\right) \cdot \mathcal{G}\left(f_{l}(\bm{s})\right)}{\left\|\mathcal{G}\left(f_{l}(\bm{x})\right)\right\| \cdot\left\|\mathcal{G}\left(f_{l}(\bm{s})\right)\right\|} ,
\end{eqnarray}
where $\bm{s}$ is the ground truth image, $\bm{x}$ is the generated image, $\mathcal{H}_{c}$ is the color histogram vector obtained in channel $c$, $\mathcal{G}$ is the Gram matrix function~\cite{gatys2015neural}, $f_l$ is the feature extraction function in the $l$-th layer of pretrained VGG-19 network~\cite{simonyan2014very}.

Except for global effects, human's perception of image style is also determined by local patterns such strokes, exquisite motifs, detailed textures, etc.
The LP metric considers both the similarity  and diversity of the local patterns between the generated image and the ground truth image.  First define $\Phi(\cdot)$ as the patch function which extract multiple sub-regions from an image. Let $ \{ \Phi_{i}^{l}(\bm{x}) \}_{i \in n_{x}}$ denote the sub-region features of image $\bm{x}$ in its $l$-th layer, and let $\{ \Phi_{j}^{l}(\bm{s}) \}_{j \in n_{s}}$ denote the sub-region features of image $\bm{s}$ in its $l$-th layer. 
Define $\mathcal{I}(i)$ as the sub-region index of image $\bm{s}$ that matches sub-region $i$ of image $\bm{x}$: $\mathcal{I}(i)=\underset{j=1, \ldots, n_{s}}{\arg \max } \frac{\Phi_{i}^{l}(\bm{x}) \cdot \Phi_{j}^{l}(\bm{s})}{\left\|\Phi_{i}^{l}(\bm{x})\right\| \cdot\left\|\Phi_{j}^{l}(\bm{s})\right\|}$.
Finally define LP as below:
\begin{eqnarray}
\label{eq:lp}
LP(\vec{x}, \vec{s})=\frac{1}{2} \cdot \frac{1}{Z} \sum_{l=1}^{N} \sum_{i=1}^{n_{x}} \frac{\Phi_{i}^{l}(\bm{x}) \cdot \Phi_{\mathcal{I}(i)}^{l}(\bm{s})}{\|\Phi_{i}^{l}(\bm{x})\| \cdot\|\Phi_{\mathcal{I}(i)}^{l}(\bm{s})\|}  \\ \nonumber
+
\frac{1}{2} \cdot \frac{1}{N} \sum_{l=1}^{N} \frac{t_{\mathcal{I}}^{l}}{t_{s}^{l}}.
\end{eqnarray}
where $Z=N \times n_{x}$,  $t_{\mathcal{I}}^{l}$ and $t_{s}^{l}$ are the numbers of different patches in $\{\Phi_{\mathcal{I}(i)}^{l}(\bm{s})\}$ and $\{\Phi_{j}^{l}(\bm{s})\}$, respectively.

\subsection{Baseline Models}

\textbf{AttnGAN~\cite{Tao18attngan}.}
AttnGAN is a representative state-of-the-art model for text-to-image generation. The vanilla GAN model is boosted by an attention-driven mechanism to guarantee fine-grained details of the image,  and a multi-stage image refinement mechanism to guarantee the visual realism of generated images. To be specific, its loss function has three parts: a unconditional loss ensuring the quality of generated images, a text-conditional loss and a image-text matching loss (DAMSM loss) forcing the representation of the text to be close to the representation of the paired image. 

In our experiment, we follow the training procedure of AttnGAN. 
The first step is to pretrain the DAMSM module which learns visually-discriminative text vectors from image-text pairs.
The hyper-parameter setting is as follows: the batch size is 48, the epoch number is 600, the word number of the text encoder is 16, default values are used for the rest hyper-parameters. 
The second step is to train both the GAN  and  DAMSM module. Different from vanilla AttnGAN, we update the generators two times and the discriminator one time at each batch because it is found that the generator is slower to converge than the discriminator. We also dynamically decrease the weight of the unconditional loss from 4, 2, to 1 and increase the weight of the DAMSM loss from 0.5, 2.5, to 5 during the first, middle and last 1/3 epochs. The goal is to ensure the model to generate images without conditioning on texts at early epochs and then learns to generate images conditioned on texts at later epochs.

\textbf{MirrorGAN~\cite{qiao2019mirrorgan}.}
MirrorGAN improves AttnGAN by proposing a text reconstruction technique to bridge the semantic gap between texts and images.  Instead of using the image-text matching loss in AttnGAN to align the representation of the texts and images, MirrorGAN uses a sequence-to-sequence model to regenerate text given the generated image, forcing the regenerate text to be close to the original one.

Similar with AttnGAN, we follow the same two-step training procedure; we also dynamically decrease the weight of the unconditional loss from 4, 2, to 1 and increase the weight of the DAMSM loss from 0.5, 2.5, to 5 during the first, middle and last 1/3 epochs.
It is interesting to compare both models to see what mechanism is the best to bridge the semantic gap between poems and paintings.

\subsection{Results}
\label{sec:results}

\begin{table*}[!htb]
\centering
\caption{The best performance of benchmark models on Zikai-Poem.}
\label{tbl:baseline-performance}
\begin{tabular}{ccccccccc}
\toprule
                           & \multicolumn{4}{c}{AttnGAN}          & \multicolumn{4}{c}{MirrorGAN}    \\  \cmidrule(lr){2-5}  \cmidrule(lr){6-9}  
Dataset                          & IS $\uparrow$        & P@1(\%) $\uparrow$  & GE $\uparrow$ & LP $\uparrow$    & IS $\uparrow$      & P@1(\%) $\uparrow$   & GE $\uparrow$ & LP $\uparrow$       \\   \midrule
CUB~\cite{wah2011caltech}                          & 4.36      & 67.82   & - & -   & 4.56       & 57.67    & - & -         \\
MS COCO~\cite{lin2014microsoft}               & 25.89      & 85.47  & - & -    & 26.47     & 74.52   & - & -          \\ 

Zikai-Poem                                                    & 2.64      & 10.00    & 0.29  & 0.31    & 3.48      & 36.00      & 0.29 & 0.33      \\

\bottomrule
\end{tabular}
\end{table*}

\begin{figure}[!htb]
    \centering
    \subfigure[Pictorial quality (IS)]{\includegraphics[width=0.4\textwidth] 
    {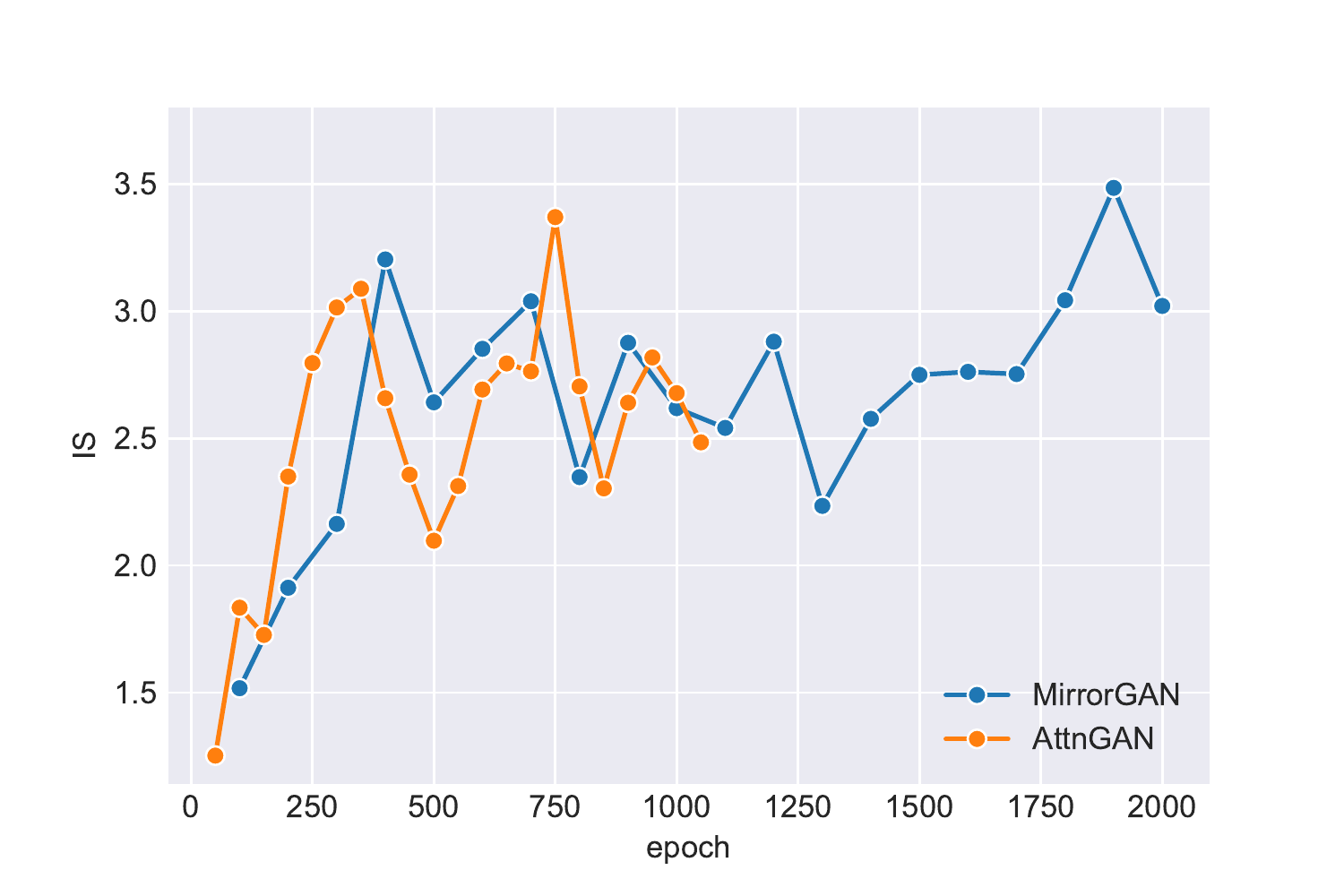}}
    \subfigure[Stylistic relevance (GE\&LP)]{\includegraphics[width=0.4\textwidth] 
    {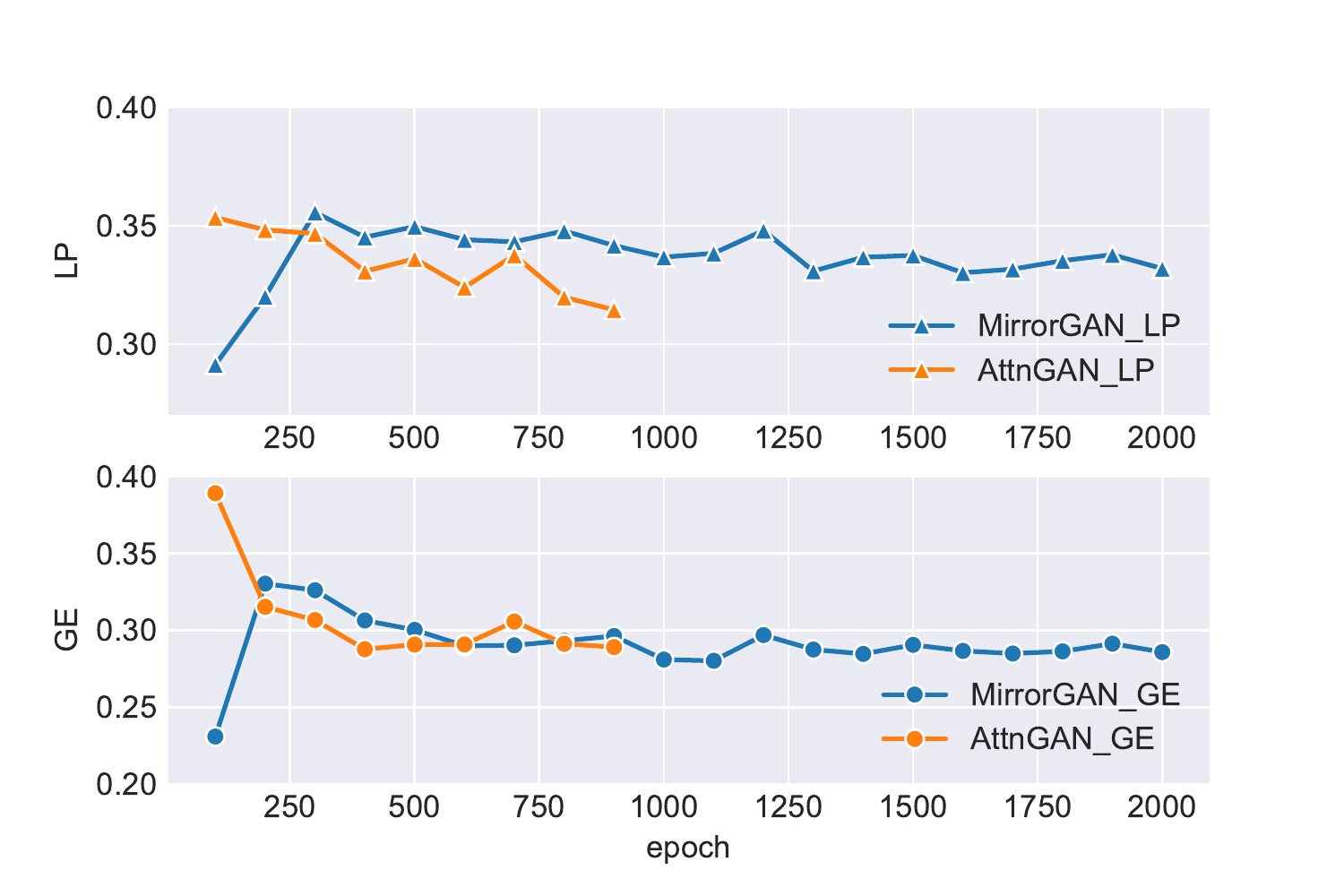}}
    \subfigure[Semantic relevance (P@1)]{\includegraphics[width=0.4\textwidth] 
    {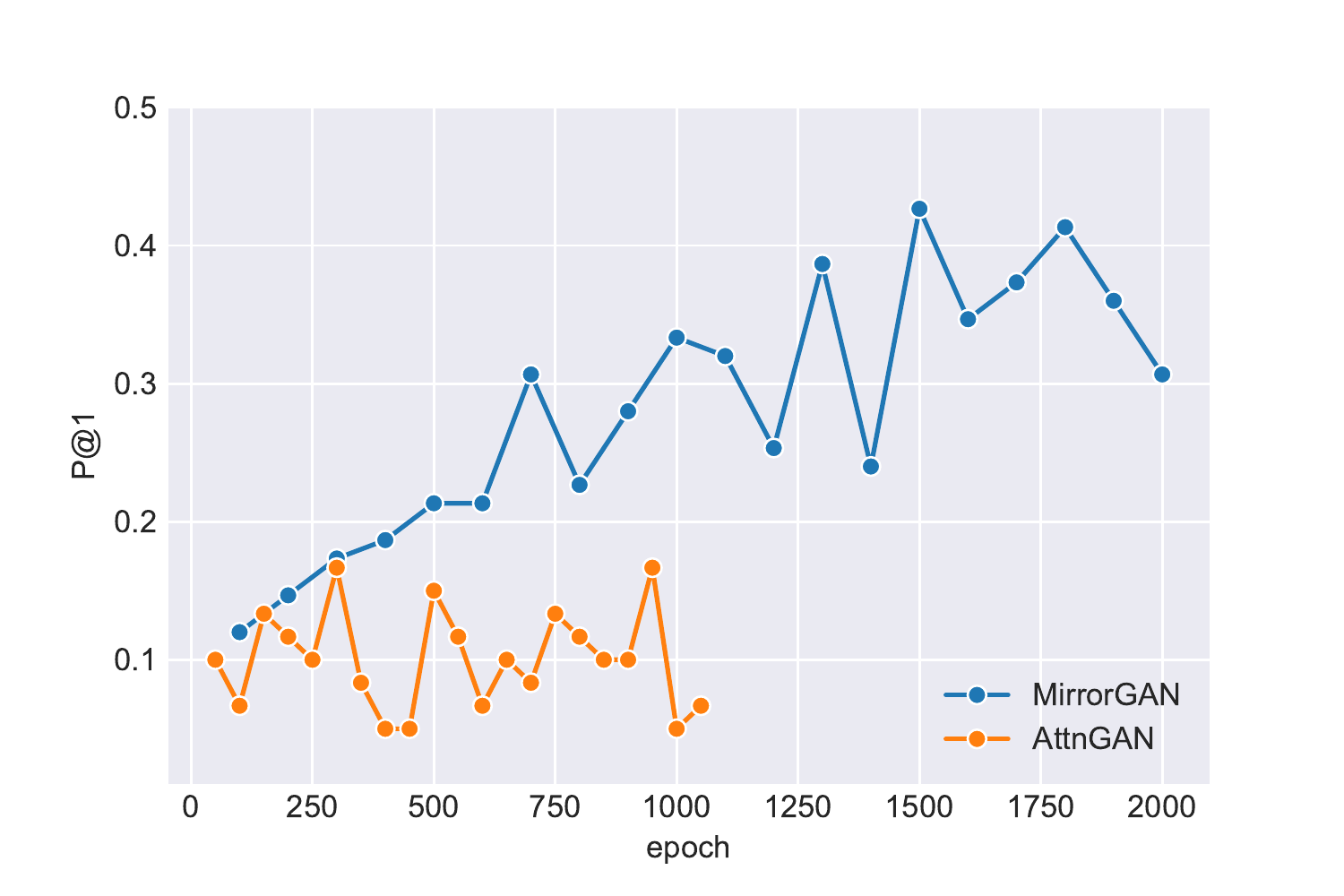}}
    \caption{Performance of benchmark models on Zikai-Poem at different training epochs.}
    \label{fig:exp-baseline-performance}
\end{figure}

\subsubsection{Performance of benchmark models on Zikai-Poem}
\label{sec:performance-benchmark-models}
\begin{figure*}[!htb]
    \centering
    \includegraphics[width=.8\textwidth]{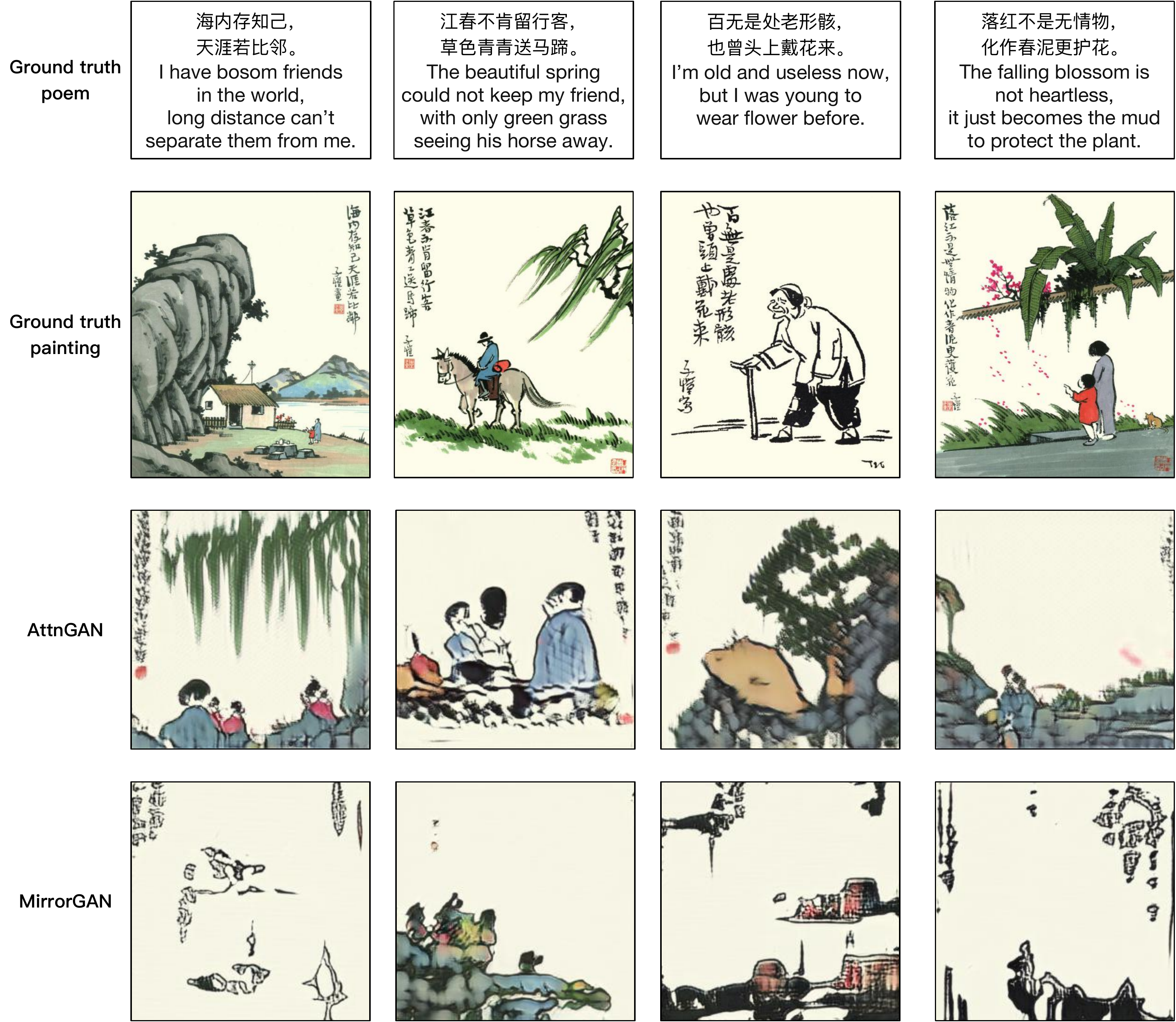}
    \caption{Example images generated by different benchmark models.}
    \label{fig:exp-benchmark-generated-images}
\end{figure*}

In this section we answer RQ1.
In Table~\ref{tbl:baseline-performance} and Fig.~\ref{fig:exp-baseline-performance}, we report the quantitative performance of benchmark models on the test set of Zikai-Poem regarding IS, P@1, GE, and LP metrics. In Fig.~\ref{fig:exp-benchmark-generated-images} we show the qualitative performance by displaying several example images generated by the benchmark models. 

\textbf{Pictorial quality.}
First, the best IS score achieved by MirrorGAN is slightly better than AttnGAN on our dataset (see Table~\ref{tbl:baseline-performance}). 
It is also similar on CUB and MS COCO. 
The IS scores on our datset and CUB are much lower than on MS COCO.
According to~\citet{salimans2016improved}, low IS score means that the label distribution $p\left(y \mid \bm{x} \right)$ and the marginal distribution $p\left(y\right)$ are close, meaning only a few class labels have none-zero probability; in other words, only a few of the 1000 objects are presented in the generated images. This is consistent with our observation: images have less types of objects  in our dataset and CUB than in MS COCO.

Second, it is interesting to see the generated images are meaningful and the objects are more or less recognisable from their outline and color, especially for AttnGAN (see Fig.~\ref{fig:exp-benchmark-generated-images}). 
For example, in the four images generated by AttnGAN in Fig.~\ref{fig:exp-benchmark-generated-images}, we can recognise objects such as willow, stone, tree, people, although it is not as delicate as those in Feng Zikai’s paintings. However, the generated images by MirrorGAN is visually worse than AtttnGAN.

Finally, we observe an inconsistency between the visual inspection of the generated images and the automatically calculated IS scores by comparing Table~\ref{tbl:baseline-performance} and Fig. \ref{fig:exp-benchmark-generated-images}. It should be noticed that IS score has limitations. According to \citet{salimans2016improved}, it is recommended to fine-tune the inception model on training datasets and to use a large number of test example (i.e. 50k) to calculate IS, whereas our test set only contains 75 examples. However, in this work we still use IS because it is a conventional metric for text-to-image generation task and has been used by most of the related works.

\textbf{Stylistic relevance.}
As mentioned before in Section~\ref{sec:eval-metrics}, humans' visual perception on image style is mainly determined by the colors and textures of images, as well as local patterns like strokes, exquisite motifs, detailed textures, etc. 
Both GE and LP scores are proposed to evaluate whether neural style transfer models can transfer the style of an image into the style of a reference image, and these neural style transfer models typically achieve high GE scores ranging from 0.6 to 0.9 and moderate LP scores ranging from 0.3 to 0.6~\citet{wang2021evaluate}. 

It is observed that the best GE and LP scores achieved by AttnGAN and MirrorGAN are very close, and the values are around 0.3 (see Table~\ref{tbl:baseline-performance}). 
It indicates a big research challenge in terms of learning style patterns like colors, textures, and strokes of the paintings in the Zikai-Poem dataset. It could be possible to integrate the merits of neural style transfer models into text-to-image generation models for future work.

\textbf{Semantic relevance.} 
First, P@1 score for MirrorGAN is higher than AttnGAN (see Table~\ref{tbl:baseline-performance}).
AttnGAN is able to select the correct text for the generated images with a chance of 13\%, and MirrorGAN with 36\%. It indicates the text-image-text mechanism of MirrorGAN plays an important role in generating semantically relevant images given texts.

Second, the P@1 score on our dataset is much lower than that on CUB and MS COCO. Furthermore, through eyeball checking of generated images, we find the generated images do not always reflect the semantic meaning of the given poems. One possible reason is that the CUB and MS COCO datasets contain multiple captions for each image and the captions have similar language structure, making it easy to learn object concepts between texts and images. Another reason could be that the training examples in both  the CUB and MS COCO datasets are large. The finding indicates our dataset is challenging in generated paintings that are semantically relevant to their paired poems.

Third, it is also an interesting observation that the Chinese calligraphic captions in the painting is also learned by the models, but the generated captions are not the correct Chinese characters. It thus poses an new challenges -- generating calligraphic captions in paintings. It could be possible to integrate Chinese calligraphic painting generation models to tackle the challenge for future work.

\begin{table}[!htbp]
\centering
\caption{Difficulty of the task perceived by human subjects (direction of arrow denotes difficulty).}
\label{tbl:exp-baseline-human-eval}
\begin{tabular}{cc}
\toprule
Score & Value \\ \midrule
Pictorial quality (1-5) $\downarrow$ & 1.48 \\
Stylistic relevance (1-5) $\downarrow$ & 1.55 \\
Semantic relevance (1-5)$\downarrow$ & 1.28  \\
Overall difficulty (1-10) $\uparrow$ & 9 \\  
\bottomrule
\end{tabular}
\end{table}


\begin{figure}[!htbp]
    \centering
    \includegraphics[width=.3\textwidth] 
    {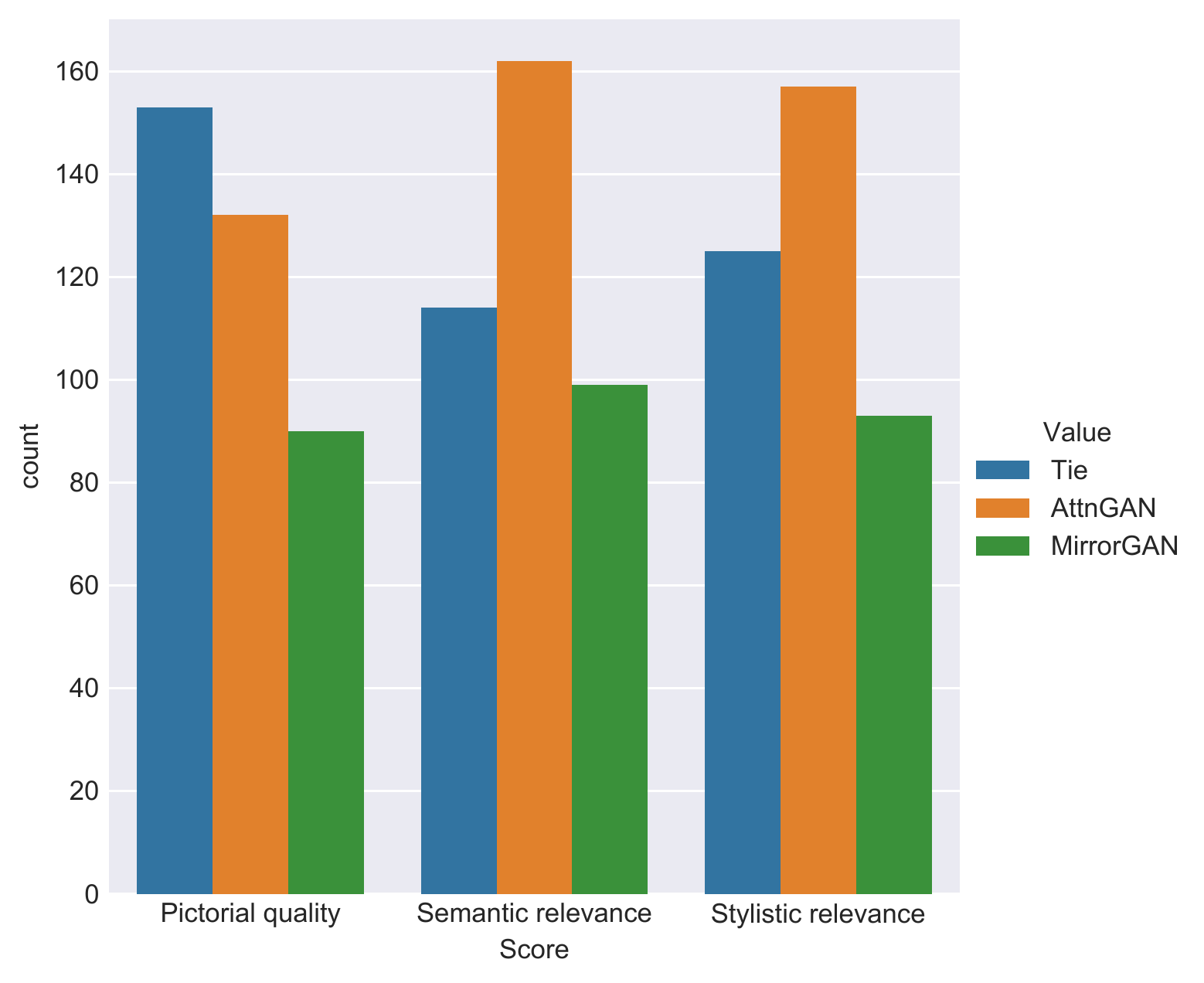}
    \caption{Results averaged over subjects for human evaluation. In the legend, \textit{Tie} means AttnGAN score is equal to MirrorGAN score, \textit{AttnGAN} means AttnGAN score is higher than MirrorGAN score, \textit{MirrorGAN} means AttnGAN score is lower than MirrorGAN score.}
    \label{fig:exp-baseline-human-eval-avg}
\end{figure}

\textbf{Human evaluation.}
In addition to automatic evaluation we also conducted human evaluation.
We are interested in understanding which of the two models -- AttnGAN and MirrorGAN perform better, and how difficult the task is for both state-of-the-art text-to-image generation models and human.

We design a questionnaire consisting of four tasks. The first one \textit{know the task} aims to make subjects familiar with the task, where they are shown five poem-painting pair examples. 
The second one \textit{choose the best generated paintings} aims to evaluate which model performs better. Subjects are shown a poem, together with two paintings -- generated by AttnGAN and MirrorGAN, and are asked to score each painting from 1 to 5 regarding pictorial quality, stylistic relevance, and semantic relevance. We use all the 75 examples in the test set.
The third one \textit{choose the best matching paintings of Feng Zikai} aims to estimate the quality of subjects in doing the task. Subjects are shown a poem, together with two Feng Zikai's paintings -- a matched one and a random one, and are asked to choose the matched painting.  We use 20 examples in the test set.
The last one \textit{post-task questionnaire} aims to collect subject profile and their feedback on the difficulty of the task of generating images from poems on a scale from 1 to 10.
We invite 5 subjects who can read and understand classical Chinese poems.

First, we calculate subject accuracy from data collected from task three, which are 90\%, 95\%, 100\%, 100\%, 100\%, respectively, indicating good reliability of them.
For each poem and painting pair in task two, we count how many times AttnGAN is higher than/equal to/lower than MirrorGAN, and denote them by \textit{Tie}/\textit{AttnGAN}/\textit{MirrorGAN}. The result is shown in 
Fig.~\ref{fig:exp-baseline-human-eval-avg}. 
We find there is a large proportion of tie, indicating the paintings generated by both models have visually similar quality. Subjects prefer AttnGAN paintings more than MirrorGAN regarding all the three criteria.
The result of human evaluation is not consist with automatic evaluation in Table~\ref{tbl:baseline-performance}, where we have analyzed the inherent problem of IS and P@1 of not reflecting the quality of generated images due to the small scale of the test set and the way IS and P@1 is calculated. 
Therefore, better automatic evaluation method are needed for our dataset in the future.

We also calculate the average scores regarding the three criteria and the difficulty value given by the five subjects to understand how difficult the task is. In Table~\ref{tbl:exp-baseline-human-eval}, we find all the three criteria are difficult to achieve as their scores are low, specifically, semantic relevance is the most difficult one, followed by pictorial quality and stylistic relevance. The subjective perception of task difficulty is 9 on a 1-10 scale. It indicates that the task is indeed challenging.

\subsubsection{Transfer learning on Zikai-Caption and TCP-Poem}
\label{sec:transfer-learning}


\begin{table}[!htb]
\centering
\caption{Performance of different transfer learning strategies on Zikai-Poem. }
\label{tbl:transfer-learning-performance}
\begin{tabular}{llcccccccc}
\toprule
  
Index & Pretrain     & Fine-tune                                          & IS  $\uparrow$   & P@1(\%) $\uparrow$  & GE $\uparrow$ & LP $\uparrow$     \\   \midrule
A &      NA            & Zikai-Poem                                     & 2.64      &10.00    & 0.29  & 0.31        \\
B & Zikai-Caption &      NA       & 2.88      & $\bm{16.67}$    & $\bm{0.54}$  & $\bm{0.34}$       \\
C & Zikai-Caption & Zikai-Poem            & $\bm{3.11}$       & 8.33    & 0.41  & 0.32      \\
D & TCP-Poem  &        NA        & 2.53      & 13.33    & 0.26  & 0.29       \\ 
E & TCP-Poem  & Zikai-Poem               & 3.07      & 11.67    & 0.35  & 0.32       \\ 
\bottomrule
\end{tabular}
\end{table} 

\begin{figure}[!htb]
    \centering
    \subfigure[Pictorial quality (IS)]{\includegraphics[width=0.4\textwidth] 
    {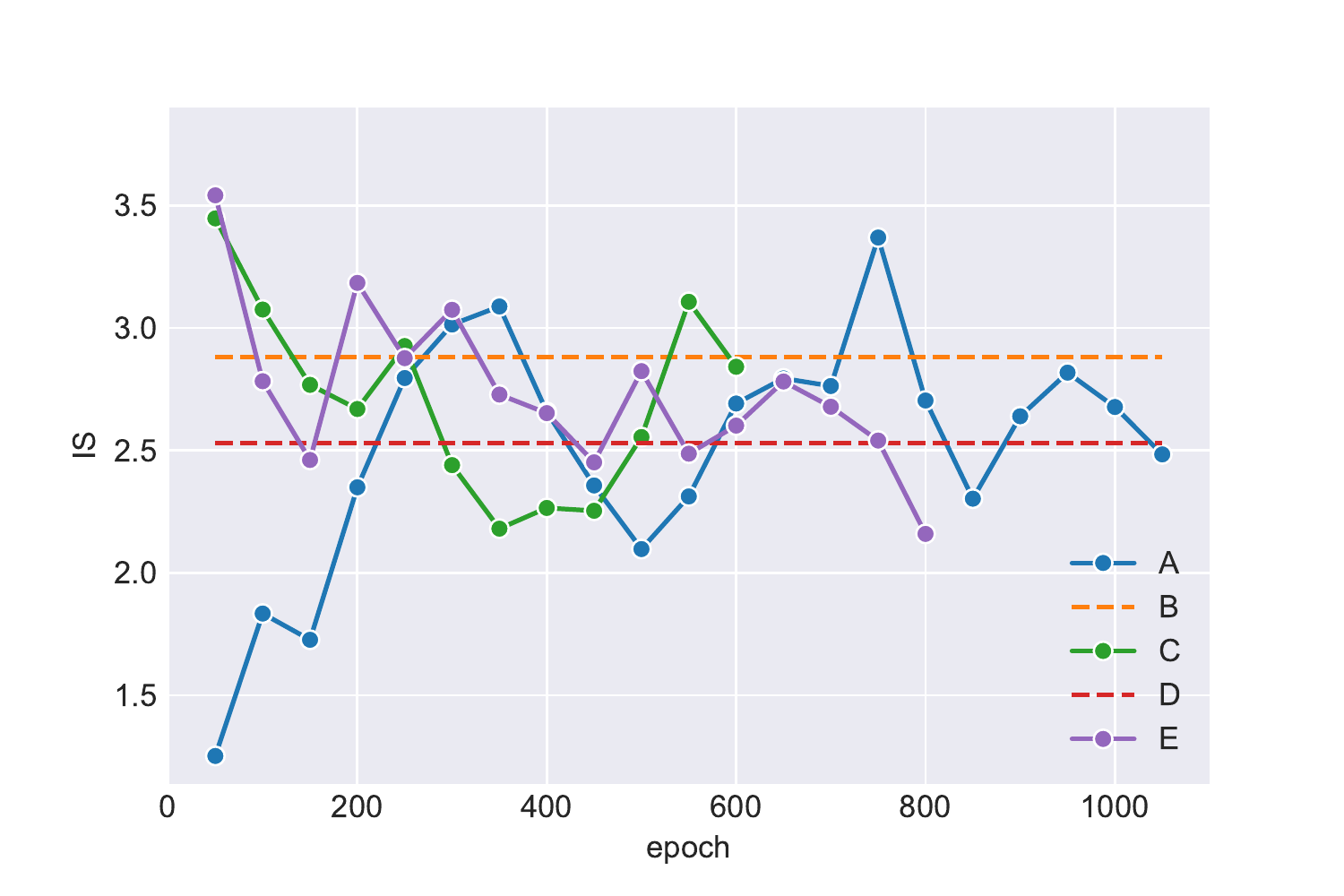}}
    \subfigure[Stylistic relevance (GE\&LP)]{\includegraphics[width=0.4\textwidth] 
    {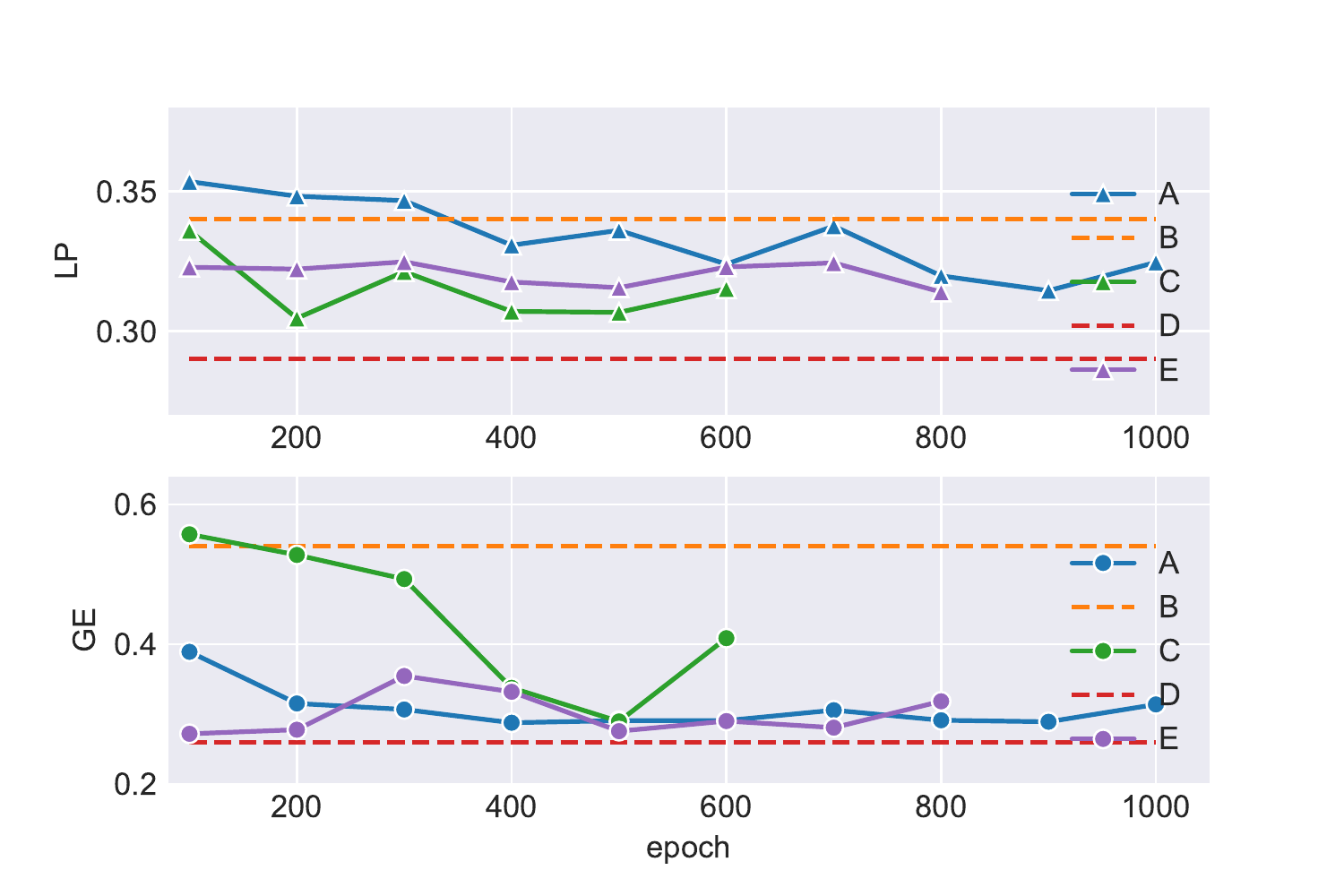}}
    \subfigure[Semantic relevance (P@1)]{\includegraphics[width=0.4\textwidth] 
    {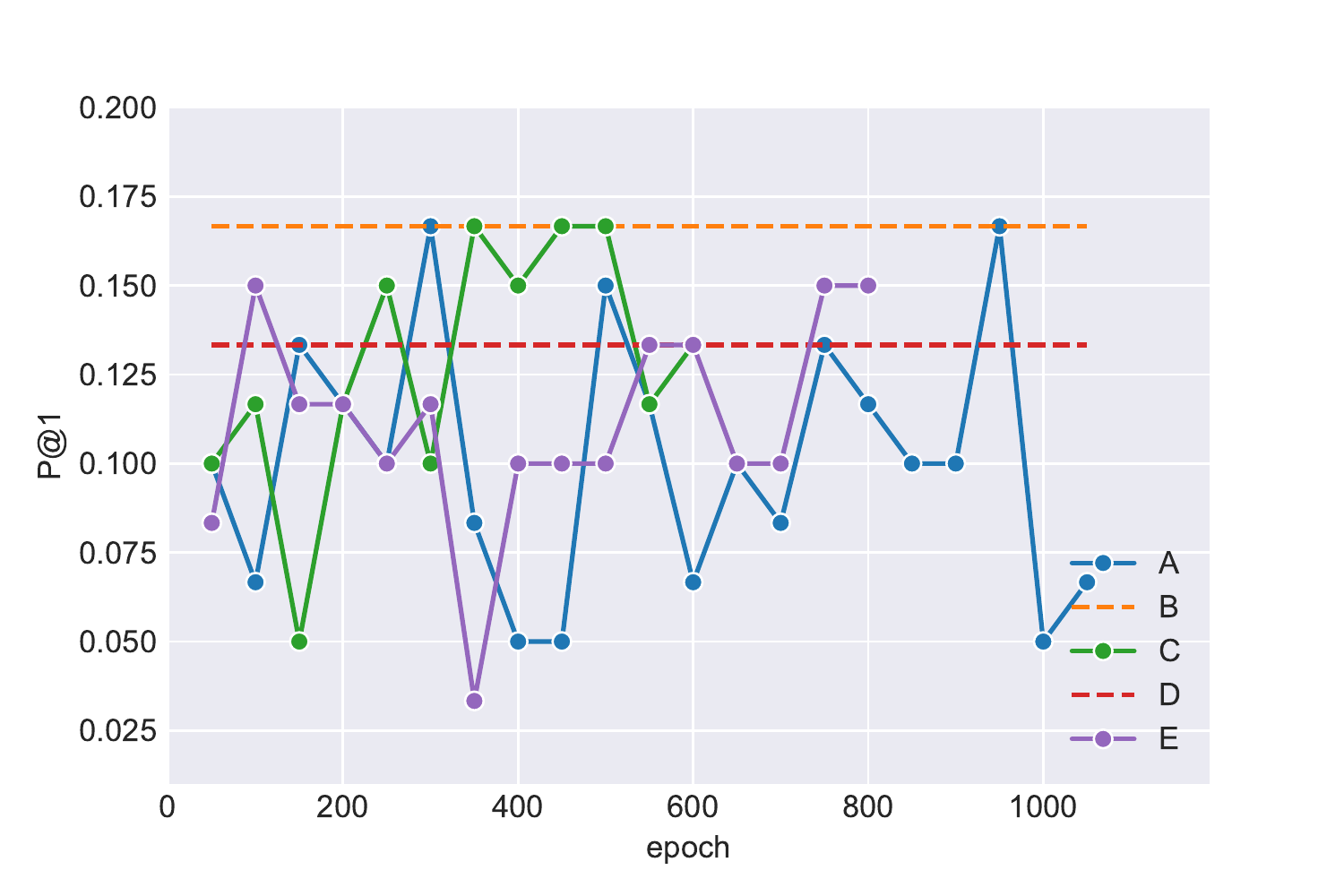}}
    \caption{Performance of different transfer learning strategies at different training epochs on Zikai-Poem.}
    \label{fig:exp-transfer-learning-performance}
\end{figure}

In this section we answer RQ2.
We study whether it is possible to apply transfer learning techniques and use the two different auxiliary datasets Zikai-Caption and TCP-Poem to improve model performance on the test set of Zikai-Poem. 
It has been shown that transferring knowledge of pre-trained networks to new domains by means of fine-tuning is a widely used practice for conditional or non-conditional GAN models, which can significantly stabilize training, shorten convergence time, and improve the quality of generated images, especially when target training data is limited~\citet{wang2018transferring, karras2020training}.
 In this work, we apply a simple pretraining-finetuning method for GAN models~\cite{wang2018transferring} -- pretraining the model on large data and finetuning both the discriminator and the generator on small data. 
 We use the AttnGAN model as it shows good performance and has been widely compared by many followed-up works. Understanding the transfer learning performance of AttnGAN can provide insight to other models like MirrorGAN as well.
 We use either Zikai-Caption or TCP-Poem for the pretraining data, and Zikai-Poem for the finetuning data.
 In total, we compare five training strategies: training on Zikai-Poem only (A), training on Zikai-Caption only (B), training on Zikai-Caption and finetuning on Zikai-Poem (C), training on TCP-Poem only (D), and training on TCP-Caption and finetuning on Zikai-Poem (E).
 The loss curves of A, B, and D show convergence with batch size=64, learning rate=2e-4 at epoch of 15, 400, and 700, respectively; the loss curves of C and E show convergence with batch size=64, learning rate=2e-4 at epoch of 300 and 200.

\begin{figure*}[!htb]
    \centering
    \includegraphics[width=.8\textwidth]{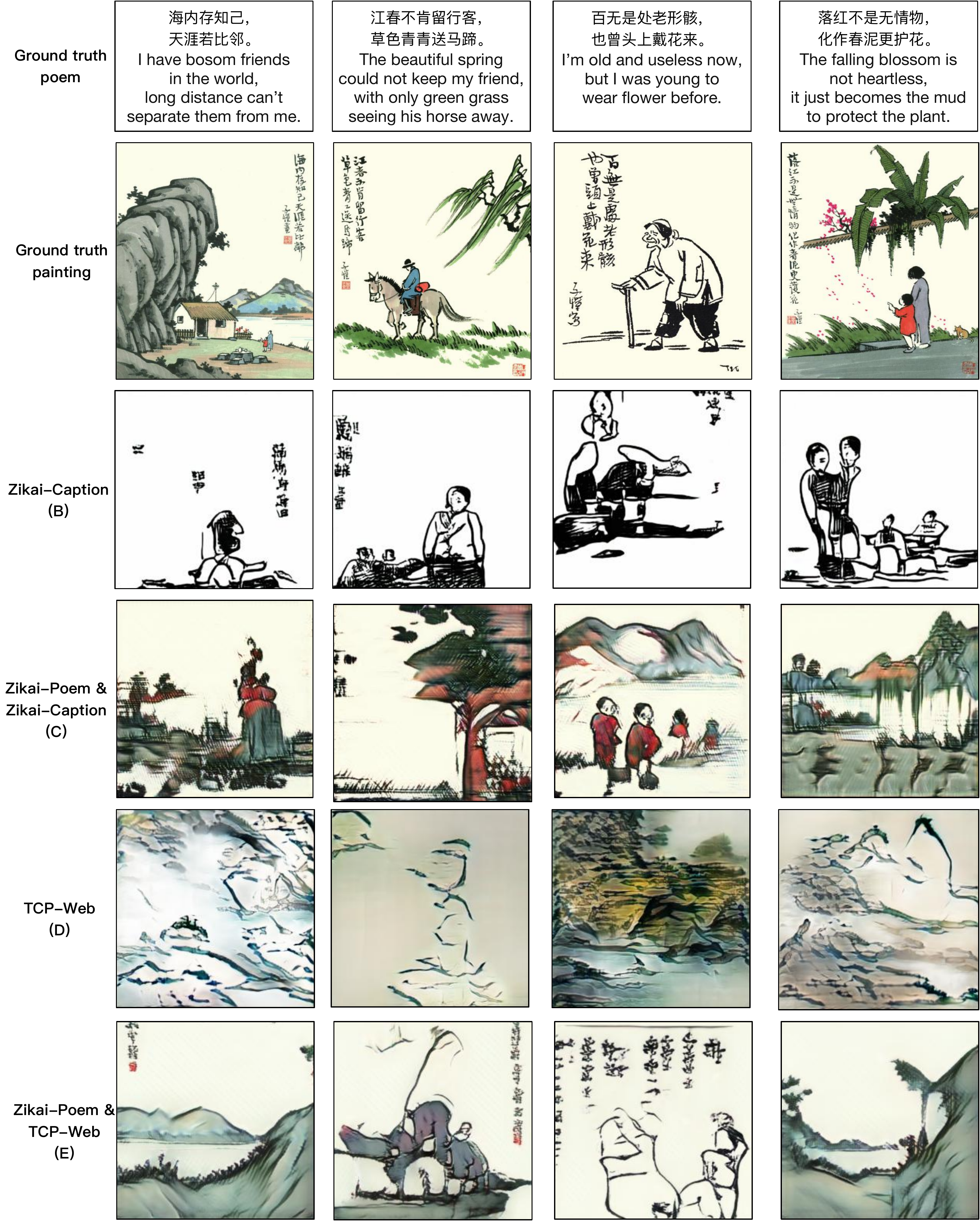}
    \caption{Example images generated by different transfer learning strategies.}
    \label{fig:exp-transfer-learning-generated-images}
\end{figure*}

In Table~\ref{tbl:transfer-learning-performance} and Fig.~\ref{fig:exp-transfer-learning-performance}, we report the  quantitative  performance  of  the five training strategies  on  the  test set  of  Zikai-Poem  by  presenting  IS,  P@1,  GE,  and  LP  metrics. In  Fig.~\ref{fig:exp-transfer-learning-generated-images} we  show  the  qualitative  performance  by  displaying several example images generated by the five training strategies.
     
Compared with the baseline model A (trained on Zikai-Poem), model B (pretrained on Zikai-Caption) performs better for all scores of IS,  P@1,  GE,  and  LP; and model C (pretrained on Zikai-Caption and finetuned on Zikai-Poem) further improves IS, however, decreases P@1, GE and LP (see Table~\ref{tbl:transfer-learning-performance}).
We also find that images generated by model B are visually similar with the training data Zikai-Caption, which are mostly monochrome and have weakly recognisable objects; while the images generated by the model C are more similar with the target data Zikai-Poem and have more recognisable objects and meaningful scenarios (see Fig.~\ref{fig:exp-transfer-learning-generated-images}). 
It indicates that the large number of paintings from Feng Zikai helps the AttnGAN model to improve pictorial quality.

Compared with the baseline model A (trained on Zikai-Poem), model D (trained on TCP-Poem) performs on par; model E (pretrained on TCP-Poem and finetuned on Zikai-Poem) improves all scores of IS,  P@1,  GE,  and  LP (see Table~\ref{tbl:transfer-learning-performance}).
In Fig.~\ref{fig:exp-transfer-learning-generated-images}), we also find that model D tend to generate images containing objects like rivers and mountains. It is because the paintings in this dataset are mostly traditional Chinese painting of rivers and mountains. Model E is able to generate images similar to Feng Zikai's style. 
However, both C and E are not able to ensure the generated images reflecting the semantics of the texts.

  


\subsubsection{Effect of model hyperparameter}

\begin{figure}[!htb]
    \centering
    \subfigure[Pictorial quality (IS)]{\includegraphics[width=0.4\textwidth]
    {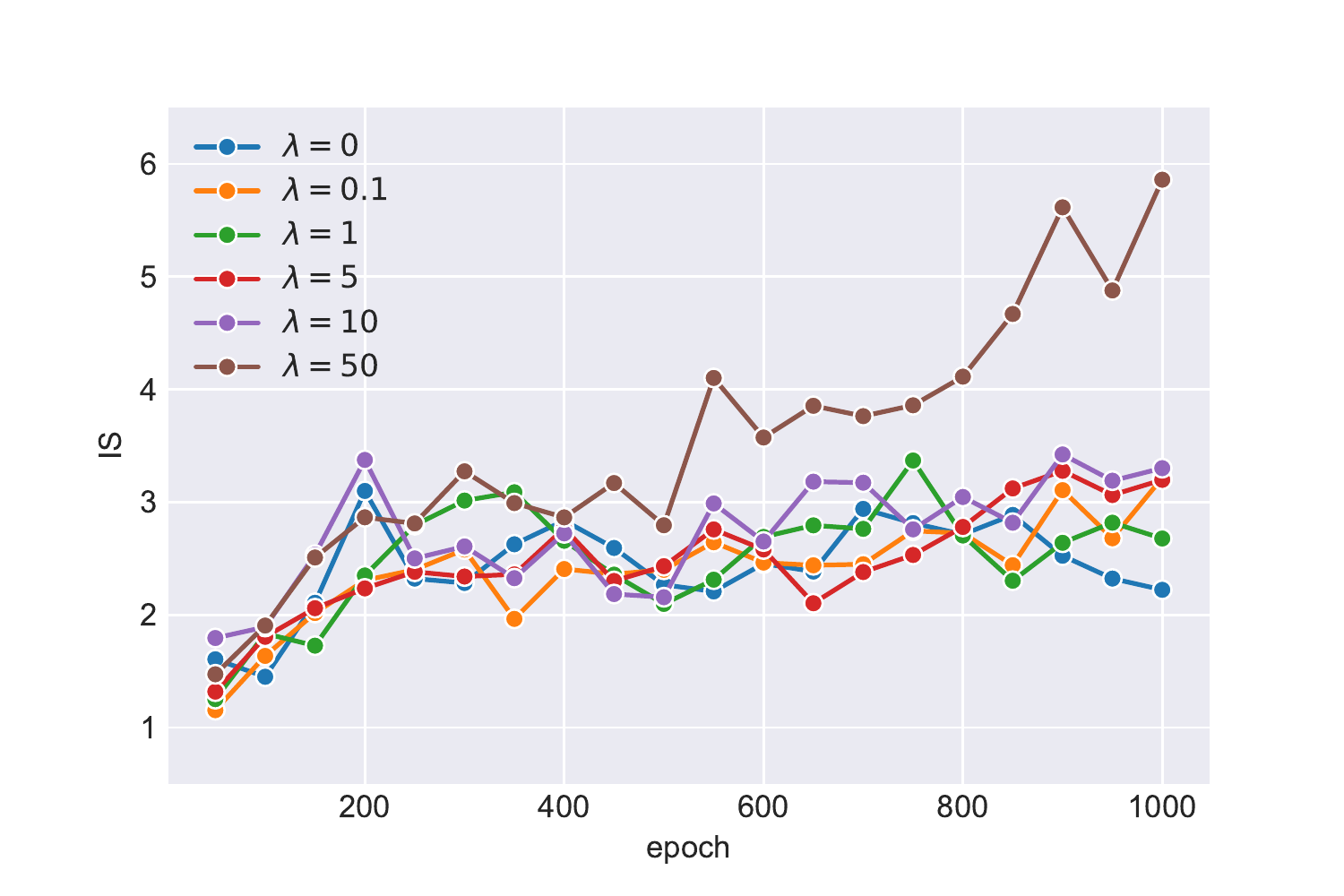}}
    \subfigure[Stylistic relevance (GE\&LP)]{\includegraphics[width=0.4\textwidth] 
    {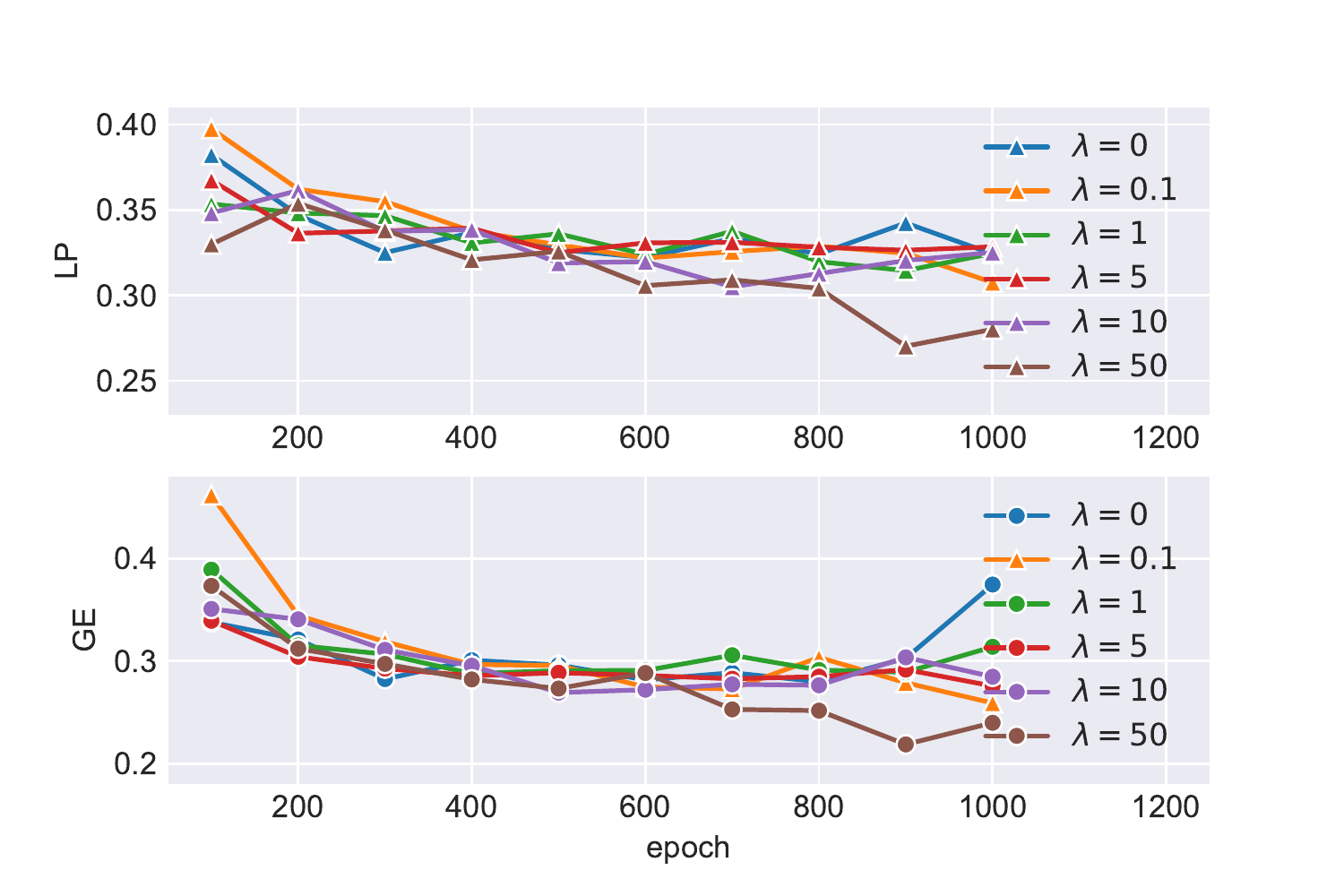}}
    \subfigure[Semantic relevance (P@1)]{\includegraphics[width=0.4\textwidth] 
    {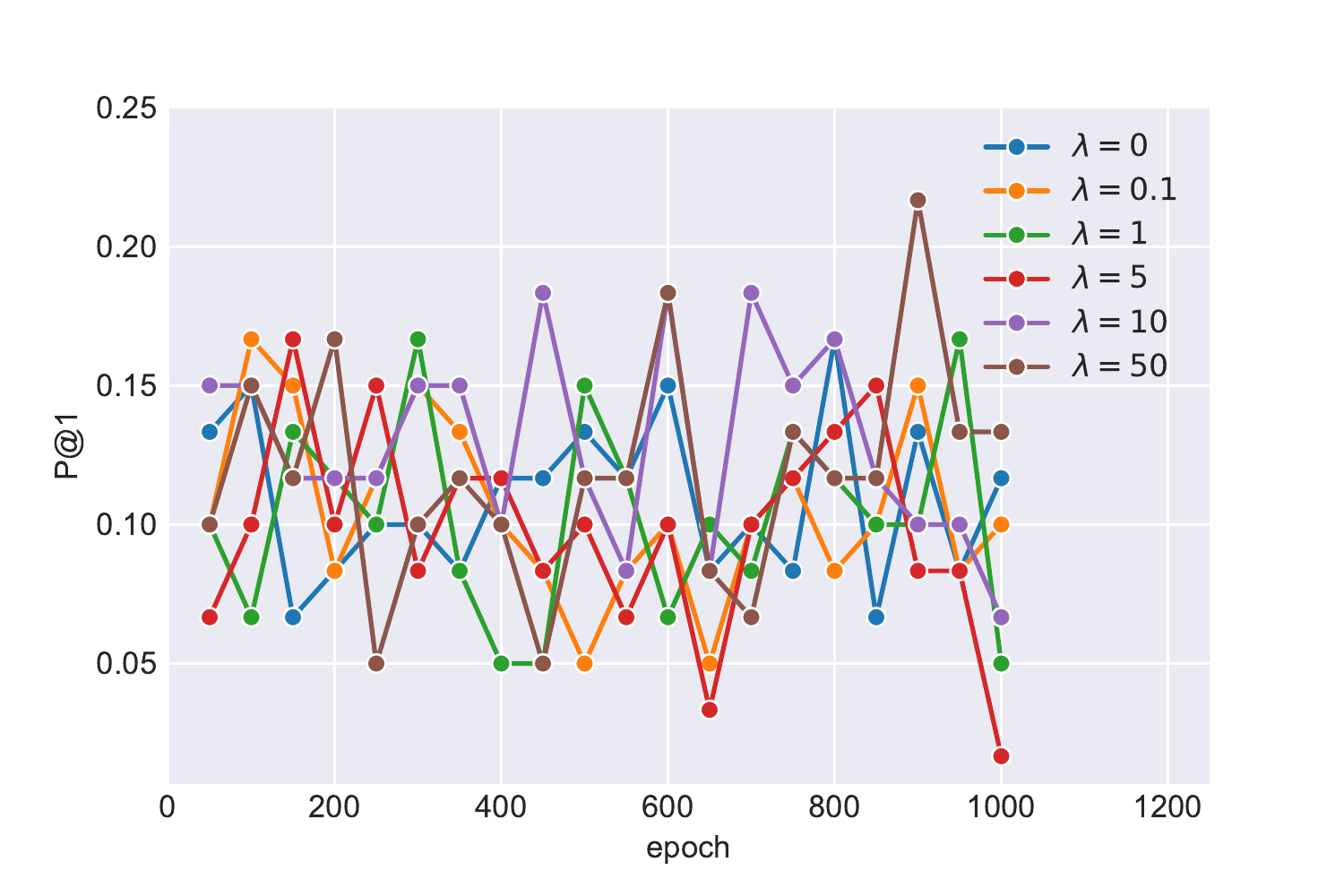}}
    \caption{Performance of different $\lambda$ of AttnGAN on Zaikai-Poem.}
    \label{fig:exp-hyperparam-metrics}
\end{figure}

The performance of the AttnGAN model is largely affected by its hyper-parameter $\lambda$~\cite{Tao18attngan}. The loss of AttnGAN is composed of the generator loss, the discriminator loss, and the DAMSM loss; and $\lambda$ is the weight of the DAMSM loss.  Proper selection of $\lambda$ helps to generate images that are better conditioned on given texts.  For example, the best $\lambda$ for CUB and MS COCO is $\lambda=5$ and $\lambda=50$~\cite{Tao18attngan}. The large $\lambda$ on MS COCO indicates that the DAMSM loss has a big impact on learning the alignment of image and text are difficult to learn. 

Following~\cite{Tao18attngan}, we search for the best $\lambda$ by increasing the value of $\lambda$ until the IS score is starting to drop on the training set of Zikai-Poem. Fig.~\ref{fig:exp-hyperparam-metrics} shows how the IS, GE\&LP, and P@1 scores of different $\lambda$ change at different training epochs. The IS score increase with $\lambda$ increasing and achieves the highest value when $\lambda=50$. The GE\&LP scores for $\lambda = 0, 1, 5, 10$ are close, but the GE\&LP scores decrease when $\lambda=50$. However, the P@1 score is not stable during training for all $\lambda$ values, indicating a challenge in learning text image alignment. 
Overall, a large $\lambda$ is needed to learn the alignment between images and texts on our dataset. 


\subsection{Research Challenges and Future Work}
\label{sec:research-challenge}

Machine learning based art and creativity continues to grow and attract a wider audience to machine learning. There have been a boost of work, for example, using generative models on new types of media creation across language and images. 
Our work is a first step towards bridging the gap between classical Chinese poems and artistic paintings.  
Both poems and paintings are creative arts. 
Although artificial intelligence has been demonstrated successful or even better than human in pattern recognition tasks regarding language and images, it is significantly lagging behind humans in creativity.
In this section we discuss several research challenges and future research directions.

A big challenge that we have not touch so far is creativity regarding paintings. As we discussed, Feng Zikai's paintings are re-creation of poems rather than literal translation of poems, his paintings bring new perspectives or thoughts that are implicitly connected to the poem.   Therefore creativity should be considered as another important criteria except for pictorial quality, stylistic relevance, semantic relevance. Yet, modelling and evaluating creativity is challenging and still at its early stage~\cite{franceschelli2021creativity}.

The challenge that we have been working on is to how to generate paintings of good pictorial quality, stylistic relevance, semantic relevance.
We have shown in Section~\ref{sec:results} that the state-of-art text-to-image generation models can generate paintings with good pictorial quality and stylistic relevance but low semantic relevance. 
However, achieving high semantic relevance is challenging due to the following characteristics of the dataset. A classical Chinese poem in our dataset is composed of multiple imageries and the paintings of Feng Zikai often only portray the most salient or emotional imageries.  Thus the poem imageries and the painting objects are not aligned in the dataset, which makes it more difficult than CUB and MS COCO.
Finally, whether a painting properly visualize a poem is often determined by the spatial relation of these imageries in the painting, rather than whether all the imageries appear in the painting.  Thus it is necessary not only to learn poem painting alignment, but also to learn spatial relation of imageries, which will largely improve painting quality.

Both the classical Chinese poems and paintings of Feng Zikai are low source data. 
It challenges the effective training of GAN models which needs large data.
Transfer learning is a possible way to tackle the problem. First, collect extra training data. In this work, we have shown how the using of the extra paintings (Zikai-Caption) and large-scale but noisy poem-painting pairs (TCP-Poem) can help improving the quality of generated paintings. More studies can be done on the two datasets. Second, use existing pretrained open-domain text-to-image models. For instance DALLE~\cite{ramesh2021zero} is a text-to-image generation model, which has a diverse set of capabilities including creating anthropomorphized versions of animals and objects, combining unrelated concepts in plausible ways, rendering text, and applying transformations to existing images. It is promising to transfer the knowledge of DALLE on our dataset to generate paintings of satisfactory pictorial quality and stylistic relevance and semantic relevance.

Another challenge is about how to better use the dataset by utilizing the rich meta data or augmenting with fine-grained annotations. 
For example, Zikai-Poem has extra information such as explanations and commentaries of poems, Zikai-Caption has an image caption for each text-image pair, which can be useful signals for poem understanding.
Labelling the imageries in poems and objects in paintings to bridge the semantic gap between poems and paintings.

\section{Related Work}

\subsection{Classical Chinese Poem}
Classical Chinese poems are easy to obtain from the web. Researchers have constructed datasets for purposes like poem generation by adding various poem annotations. For example, 
\citet{liu2018images2poem} released 128k classical Chinese poems (quatrain) augmented with automatically-extracted keywords for poem generation. 
\cite{chensentiment:19} manually labelled classical Chinese poems with sentiment labels, which can be used for poem sentiment detection or sentiment-aware poem generation.
\cite{jiuge:19} constructed a poem dataset with rhythm and rhyme annotations. 
Such datasets enhance the understanding of classical Chinese poems in various perspectives. In this paper, we propose to provide novel understanding by generating paintings of different artistic styles for poems. 

There are also work on the opposite direction with this work -- image inspired poem generation \cite{xu2018images, liu2018beyond, cheng2018image, liu2018multi, zhang2014chinese, liu2018images2poem}, where photo-poem pairs are collected form the web. The datasets are different from ours, but the automatic method to collect photo-poem pairs is inspiring.  Moreover, the multi-modal poem generation also inspires the modelling of poem-to-painting generation such as poem representation learning and poem-image aligned representation learning.

\subsection{Text-to-Image Generation}
Text-to-image generation refers to generating visually realistic images that match given text descriptions. Existing datasets have benefited the development of text-to-image generation models.
The CUB dataset~\cite{wah2011caltech} is originally designed for bird classification. It consists of 200 categories of bird images with textual captions. The dataset is relatively simple for the text-to-image generation task because the images are only about birds and the captions are based on templates with limited bird attributes.
Another similar datasets is the Oxford-102 flower image dataset~\cite{nilsback2008automated}.
The MS COCO dataset~\cite{lin2014microsoft} is more difficult in the sense that it contains diverse categories (around 90) of open-domain objects. 
The existing datasets for training  text-to-image models are not perfect to train poem-to-painting models due to the different domains of texts and different styles of images. Therefore we constructed a new dataset in this paper.

State-of-the-art text-to-image generation models are based on GAN.
GAN consists of a generator that learns to generate new data from the training data distribution and a discriminator that learns to identify the generated data from the real data. 
In text-to-image generation models, the image generator is conditioned on  text vectors transformed from the text description. The learning of the shared semantics of paired text and image are key to text-to-image generation, which are learned through attention mechanism~\cite{Tao18attngan}, text and image similarity loss~\cite{qiao2019mirrorgan} etc.

%
%

\section{Conclusion}

In this work we propose a new task -- artistic visualization of classical Chinese poems, where the goal is to generate images of a certain artistic style for classical Chinese poems. 
For this purpose, we construct a new dataset called Paint4Poem. 
The first part consists of 301 high-quality poem-painting pairs collected manually from Feng Zikai's paintings, which serves as both the training and test set.
As its small scale poses a big challenge for training, we introduce the second part, which consists of 3,648  caption-painting pairs collected manually from Feng Zikai's paintings and 89,204 poem-painting pairs collected automatically from the web. We expect the former to help learning the artist painting style as it almost contains all his paintings, and the latter to help learning text image alignment. 
We analyze the Paint4Poem dataset in three aspects: poem diversity, painting style, and the semantic relevance between paired poems and paintings. 
We create a benchmark for the dataset: we train two state-of-the-art text-to-image generation models -- AttnGAN and MirrorGAN, and evaluate their performance in terms of image pictorial quality, image stylistic relevance, and semantic relevance between images and poems. 
The results indicate that it is able to generate paintings that have good pictorial quality and mimic Feng Zikai's style, but the reflection of the semantics of given poems is limited.
The dataset also poses many interesting research directions on this task, including transfer learning, few-shot learning, text-to-image generation for low-resource data etc.

\ifCLASSOPTIONcaptionsoff
  \newpage
\fi



%

\bibliography{reference}
\bibliographystyle{IEEEtran}

%







\end{document}